\documentclass[lettersize,journal]{IEEEtran}
\usepackage{amsmath,amsfonts}
\usepackage{algorithmic}
\usepackage{array}
\usepackage[caption=false,font=normalsize]{subfig}
\usepackage{textcomp}
\usepackage{stfloats}
\usepackage{url}
\usepackage{verbatim}
\usepackage{graphicx}
\usepackage[table,dvipsnames]{xcolor} %
\usepackage{cite}
\def\BibTeX{{\rm B\kern-.05em{\sc i\kern-.025em b}\kern-.08em
    T\kern-.1667em\lower.7ex\hbox{E}\kern-.125emX}}
\usepackage{balance}

\usepackage{algorithm}
\usepackage{hyperref}
\usepackage{url}
\usepackage{booktabs}
\usepackage{wrapfig}
\usepackage{etoc}
\usepackage{lipsum}
\usepackage{float}
\usepackage{makecell}
\usepackage{array}
\usepackage{multirow}
\usepackage{threeparttable}
\usepackage{siunitx}
\usepackage{marvosym}
\usepackage{etoolbox}
\usepackage{nicematrix}
\usepackage{etoc}
\usepackage{titletoc}
\usepackage{booktabs,tabularx}
\usepackage{enumitem}
\setlist{nosep}
\definecolor{darkgreen}{rgb}{0.0, 0.75, 0.0}
\definecolor{lightgreen}{rgb}{0.4, 0.4, 0.4}
\newcommand{\gdelta}[1]{{\footnotesize\textcolor{darkgreen}{#1}}}
\usepackage{placeins}
\usepackage{adjustbox}
\usepackage{tabularx}
\usepackage{stfloats}

\begin{document}

\title{MemoryVLA++: Temporal Modeling via Memory and Imagination in Vision-Language-Action Models}

\author{

Hao Shi$^\dagger$, Weiye Li, Bin Xie$^\dagger$, Yulin Wang, Renping Zhou, Tiancai Wang, Xiangyu Zhang, \\
Ping Luo,~\IEEEmembership{Senior Member,~IEEE,}, 
Gao Huang~\IEEEmembership{Member,~IEEE,}$^\S$

\thanks{Hao Shi is with the Department of Automation, BNRist, Tsinghua University, Beijing, China and also with the Department of Computer Science, The University of Hong Kong, Hong Kong, China.}
\thanks{Weiye Li, Yulin Wang, Renping Zhou, and Gao Huang are with the Department of Automation, BNRist, Tsinghua University, Beijing, China.}
\thanks{Bin Xie and Tiancai Wang are with Dexmal, Beijing, China.}
\thanks{Xiangyu Zhang is with StepFun, Beijing, China.}
\thanks{Ping Luo is with the Department of Computer Science, The University of Hong Kong, Hong Kong, China.}

\thanks{$^\dagger$ denotes project leaders, and $^\S$ denotes corresponding author.}
\thanks{E-mail: shi-h23@mails.tsinghua.edu.cn; gaohuang@tsinghua.edu.cn.}
\thanks{Project page: \url{https://shihao1895.github.io/MemoryVLA-PP-Web}.}
}

\maketitle

\begin{abstract}
Temporal modeling is essential for robotic manipulation, as effective control requires both memory of past interactions and imagination of future states. However, most VLA models rely primarily on the current observation and therefore struggle with long-horizon, temporally dependent tasks. 
Cognitive science suggests that humans rely on working memory to buffer short-lived context, the hippocampal system to preserve episodic memory of past experience, and internal models to imagine possible future state evolution. 
Inspired by these mechanisms, we propose MemoryVLA++, a full temporal modeling framework that equips VLA models with memory and imagination for robotic manipulation. 
A pretrained VLM encodes the current observation into perceptual and cognitive tokens, forming working memory. These tokens query a Perceptual-Cognitive Memory Bank to retrieve relevant historical context. This bank stores low-level details and high-level semantics from past interactions, and is updated through redundancy-aware consolidation. 
A world model imagines future states in a denoising latent space, and the imagined latents are integrated under memory guidance to form full temporal-aware tokens. 
The resulting tokens condition a diffusion action expert to predict temporally consistent action sequences. 
We conduct extensive experiments on 5 simulation benchmarks and 3 categories of real-robot tasks across 3 robots, covering general manipulation, long-horizon temporal tasks, robustness, and generalization. 
Our method achieves strong performance across Libero, SimplerEnv, Mikasa-Robo, Calvin, Libero-Plus, and diverse real-robot tasks, validating the effectiveness of full temporal modeling with memory and imagination. 
For example, on real robots, it achieves +9\% gains on general manipulation, +26\% on long-horizon memory-dependent tasks, and +28\% on long-horizon imagination-dependent tasks. 
\end{abstract}


\section{Introduction}
\label{sec:intro}

\begin{figure*}[t]
    \centering
    \includegraphics[width=1.0\linewidth]{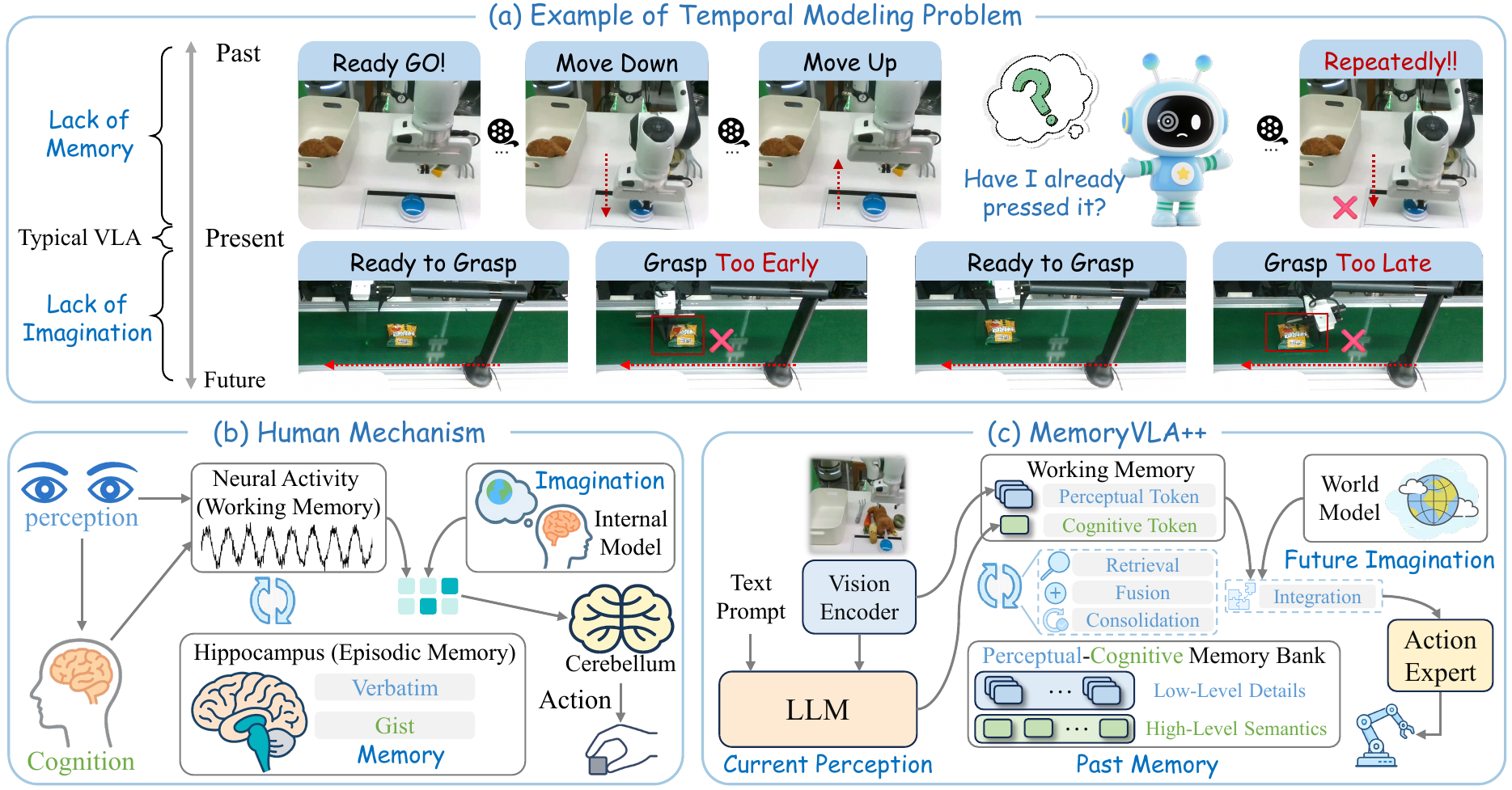}
    \caption{
    Motivation of MemoryVLA++. 
    (a) Button pressing shows the need for memory: similar visual states before and after pressing make it hard to know whether the button has already been pressed. Dynamic-conveyor grasping shows the need for imagination: predicting future object motion helps grasp at the right time.
    (b) Humans leverage the hippocampal system to maintain working-episodic memory, and use internal models to imagine future state evolution. 
    (c) Inspired by these, MemoryVLA++ enables full temporal modeling in VLA models by combining present perception, past memory, and future imagination.
    }
    \label{fig:motivation}
\end{figure*}

\IEEEPARstart{V}{ision-Language-Action} (VLA) models~\cite{kim2025openvla,black2025pi_0,li2024cogact}, powered by large-scale cross-embodiment robotic datasets~\cite{o2024open,brohan2023rt,khazatsky2024droid,bu2025agibot} and pretrained Vision-Language Models (VLMs)~\cite{karamcheti2024prismatic,bai2023qwen}, have achieved remarkable progress in robotic manipulation. 
However, typical VLA models such as OpenVLA~\cite{kim2025openvla} and $\pi_0$~\cite{black2025pi_0} rely solely on the current observation, thereby overlooking temporal dependencies and performing poorly on long-horizon temporal tasks. 
As shown in Fig.~\ref{fig:motivation}(a), \textit{Button Pressing} tasks require memory, since the observations before and after pressing exhibit almost no visual difference, making it difficult to determine whether the button has already been pressed. 
Meanwhile, \textit{Dynamic-Conveyor Grasping} tasks require imagination, as perceiving how objects will move on a dynamic conveyor enables the robot to grasp at a more appropriate time. 
These examples highlight that effective robotic manipulation requires both remembering past interactions and anticipating how the scene will evolve in the future. 

A straightforward way is to expand the VLA input with additional frame-level observations from both the past and the future. Specifically, the most recent $N$ history frames~\cite{liu2026ttf,jang2025contextvla,wang2025lola} and predicted future video frames~\cite{zhang2026foreact,long2026scaling,du2023learning,black2024zero} are fed into the VLA model. 
However, simply expanding the visual observation sequence is inefficient for robotic manipulation. 
On the past side, concatenating consecutive history frames incurs quadratic self-attention cost and introduces substantial temporal redundancy, limiting the usable context length and impairing long-term dependency modeling. This issue is especially pronounced in robotic manipulation, where slow motions stretch decision-relevant interactions over longer time spans. 
On the future side, future video prediction is computationally expensive and often focuses on pixel-level fidelity rather than control-relevant dynamics. Directly feeding predicted frames into the VLA model may further propagate visual prediction errors to action generation. 
This motivates a temporal modeling framework that efficiently compresses past experience into long-term memory and captures future dynamics through compact, decision-relevant imagination. 

Research in cognitive science~\cite{baddeley1974working,tulving1972episodic,reyna1995fuzzy,craik1967nature,grush2004emulation} suggests that humans handle manipulation tasks by using working-episodic memory to retain past experiences and internal models to anticipate future state evolution, as illustrated in Fig.~\ref{fig:motivation}~(b). 
Multimodal sensory information is encoded into perceptual and cognitive representations, which are temporarily maintained in working memory for immediate decision-making. 
 In parallel, episodic memory, a long-term memory system closely associated with the hippocampus, stores past experiences as both verbatim representations that preserve precise details and gist representations that capture abstract semantics. 
Beyond memory, internal models further enable humans to anticipate possible future states before action execution. 
During execution, working memory retrieves decision-relevant contexts from episodic memory and integrates them with current representations and future anticipation, forming a temporal representation that guides motor execution through the cerebellum, while new experiences are consolidated into episodic memory. 

\begin{figure*}[t]
    \centering
    \includegraphics[width=0.85\linewidth]{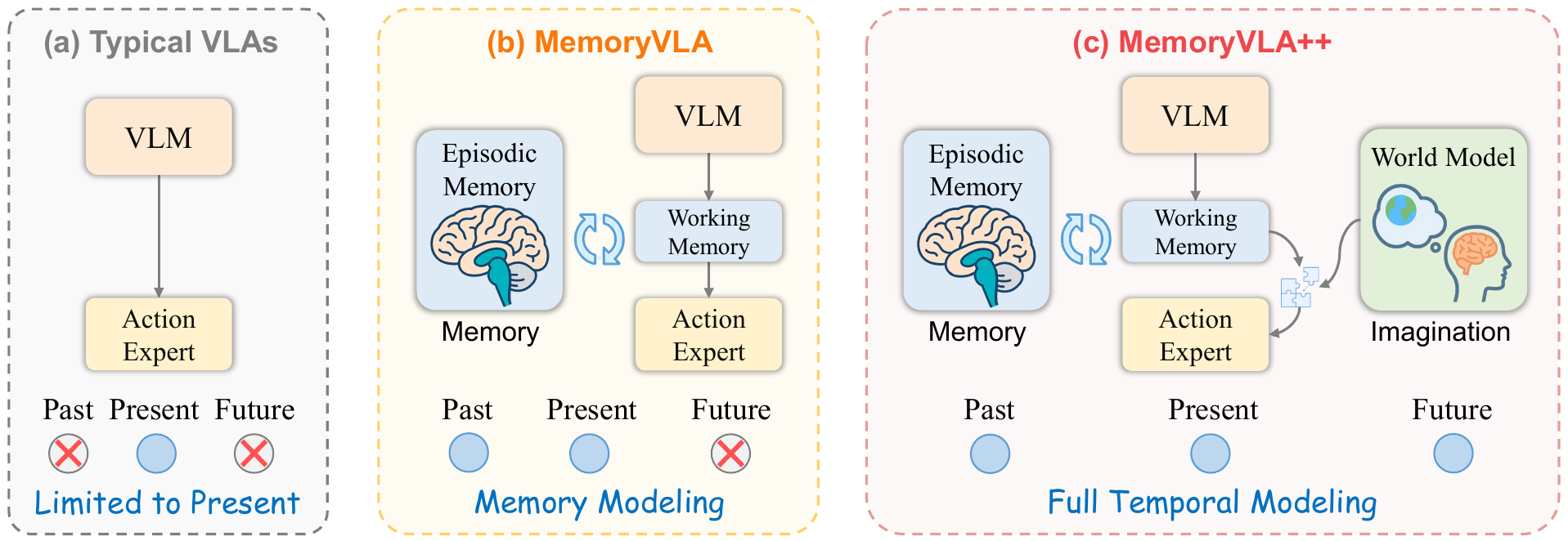}
    \caption{Comparison of the main ideas of three paradigms: Typical VLAs are reactive and rely only on the present observation, MemoryVLA introduces a working memory-episodic memory mechanism to capture past temporal dependencies, and MemoryVLA++ further extends this by incorporating future imagination via a world model for full temporal modeling.}
    \label{fig:idea}
\end{figure*}

Drawing on cognitive science insights, we propose MemoryVLA++, a full temporal modeling framework that equips VLA models with memory and imagination for robotic manipulation, as shown in Fig.~\ref{fig:motivation}~(c).  
A vision encoder extracts perceptual tokens from the current observation, while an LLM reasons over these tokens and language tokens to produce cognitive tokens using its commonsense prior. Perceptual and cognitive tokens jointly form the working memory. 
To capture long-term temporal context, a Perceptual-Cognitive Memory Bank (PCMB) stores low-level perceptual details and high-level cognitive semantics from past interactions. 
Working memory queries the PCMB to retrieve decision-relevant historical contexts, which are adaptively fused with current tokens through a gating mechanism. Meanwhile, the PCMB is updated through redundancy-aware consolidation, merging temporally adjacent and semantically similar entries to keep memory compact. 
To anticipate future state evolution, a video-generation world model performs partial denoising in the latent space to obtain imagined future tokens. 
Guided by memory-augmented tokens, these imagined tokens are integrated into full temporal tokens that combine present perception, past memory, and future imagination. 
The resulting tokens condition a diffusion action expert, where cognitive tokens provide high-level semantic guidance and perceptual tokens supply fine-grained visual details, producing temporally consistent robotic actions. 
Fig.~\ref{fig:idea} compares the main idea of MemoryVLA++ with typical VLAs and MemoryVLA, illustrating full temporal modeling with memory and imagination.

We conduct extensive experiments on 5 simulation benchmarks and 3 categories of real-robot tasks across 3 robots, covering nearly 200 tasks with diverse variations. 
In simulation, MemoryVLA++ achieves success rates of 98.4\% on Libero~\cite{liu2023libero} and 74.0\% on SimplerEnv~\cite{li2025evaluating}, consistently outperforming strong baselines with a maximum gain of 16.7 percentage points on SimplerEnv. 
On long-horizon and temporal tasks, MemoryVLA++ achieves a 44.4\% success rate on Mikasa-Robo~\cite{cherepanov2025memory} and a score of 4.29 on Calvin~\cite{mees2022calvin}, improving over the baseline by 15.0 percentage points on Mikasa-Robo. 
For robustness and generalization, MemoryVLA++ further achieves an 82.7\% success rate on Libero-Plus under diverse task and environment variations. 
In real-robot experiments, we evaluate our method on general manipulation, long-horizon memory-dependent tasks, and long-horizon imagination-dependent tasks. It achieves scores of 85\%, 83\%, and 77\%, outperforming the baseline by 9, 26, and 28 percentage points, respectively. 
These results validate the effectiveness of full temporal modeling with memory and imagination for robotic manipulation. 

This work substantially extends the conference version MemoryVLA~\cite{shi2025memoryvla} with several important new contributions, as summarized below.

\begin{itemize}
    \item We advance VLA temporal modeling from past-only memory to full temporal modeling over the past, present, and future, equipping VLA models with both memory and imagination (Fig.~\ref{fig:idea}). Inspired by cognitive science, the framework extends the previous perceptual-cognitive memory mechanism with a world-model-based imagination mechanism, forming MemoryVLA++ (Fig.~\ref{fig:overall}).
    
    \item To leverage the world model for future imagination, we introduce two key components: (1) a latent-space imagination generation that uses the world model to capture decision-relevant future dynamics via partial denoising, avoiding costly pixel-level prediction; (2) a memory-guided imagination integration module that adaptively integrates imagined future latents with memory-augmented tokens to produce full temporal-aware representations. (Sec.~\ref{sec:imag}, Fig.~\ref{fig:imag_details}).
    
    \item The experimental evaluation has been substantially expanded. We report updated MemoryVLA++ results on five simulation benchmarks, including Libero, SimplerEnv, Mikasa-Robo, Calvin, and Libero-Plus, covering nearly 200 tasks with diverse variations (Tabs.~\ref{tab:libero}, \ref{tab:simpler}, \ref{tab:mikasa}, \ref{tab:calvin}, and~\ref{tab:libero_plus}). Among them, Calvin and Libero-Plus are newly added to evaluate long-horizon performance and robustness \& generalization, respectively. 
    
    \item We further expand the real-robot evaluation with long-horizon imagination-dependent tasks, demonstrating MemoryVLA++'s effectiveness in practical deployment (Tab.~\ref{tab:real}). The evaluation is also extended to a dual-arm ARX5 platform, in addition to the previously used single-arm robots, covering three robots in total (Fig.~\ref{fig:robot}).
    
    \item Additional analytical results are provided, including ablation studies of new components (Tab.~\ref{tab:ablation_imag}), analysis of efficiency (Tab.~\ref{tab:efficiency}), analysis of the world model (Fig.~\ref{fig:vis_world}, Tab.~\ref{tab:world_model}), exploration of stronger Qwen~\cite{bai2023qwen} VLM backbones with dexbotic~\cite{xie2025dexbotic} pretraining (Tab.~\ref{tab:backbone}), and new visualization results for real-robot tasks (Fig.~\ref{fig:vis_real}).
\end{itemize}

\section{Related Work}

\subsection{Vision-Language-Action Models}
Driven by advances in visual foundation models~\cite{oquab2024dinov2,zhai2023sigmoid,zheng2024denseg,zheng2025densegrounding,zhang2025grounding,wang2025emulating,huang2022glance,wang2024uni}, robot imitation learning has progressed rapidly, yet remains confined to small, task-specific policies with limited generalization~\cite{chi2023diffusion,zhao2023learning,goyal2023rvt,shi2026spatialactor}. 
To overcome these limitations, the success of VLMs~\cite{achiam2023gpt,touvron2023llama,bai2023qwen} and large-scale robot datasets, such as OXE~\cite{o2024open} and Agibot~\cite{bu2025agibot}, has spawned the vision-language-action (VLA) paradigm~\cite{kim2025openvla,black2025pi_0,liu2026hybridvla,yue2024deer,sun2025geovla}. 
RT-2~\cite{zitkovich2023rt} and OpenVLA~\cite{kim2025openvla} tokenize continuous actions into discrete tokens and use VLMs for autoregressive prediction as if generating language. 
In contrast, $\pi_0$~\cite{black2025pi_0}, CogACT~\cite{li2024cogact}, and GR00T-N1~\cite{bjorck2025gr00t} adopt diffusion-based policies~\cite{chi2023diffusion} as action heads, leveraging iterative denoising to sample continuous control trajectories and capture diverse multimodal behaviors. 
However, most existing VLA models mainly rely on current observations, without explicitly modeling past interaction history or future state evolution, and therefore struggle with long-horizon temporal tasks. 

\subsection{Memory Modeling for Robotic Manipulation}
Memory modeling has been extensively studied in computer vision and autonomous driving~\cite{wang2023exploring,feng2023open}, yet it has not been fully explored in robotic manipulation. 
Octo~\cite{team2023octo}, RoboVLMs~\cite{li2026matters}, and Interleave-VLA~\cite{fan2025interleave} organize past observations into interleaved image-text sequences, while TTF-VLA~\cite{liu2026ttf} and ContextVLA~\cite{jang2025contextvla} directly concatenate neighboring frames. Although these approaches provide some temporal context, they are computationally expensive and limited to short horizons. 
Another line of work encodes history into latent representations. CronusVLA~\cite{li2025cronusvla}, 4D-VLA~\cite{zhang20264d}, and HAMLET~\cite{koo2026hamlet} concatenate neighboring latent frames, SAM2Act~\cite{fang2025sam2act} augments latent features with heatmap masks, and HiF-VLA~\cite{lin2026hif} captures dynamics through motion cues. Although these methods use historical information efficiently, they rely on sliding windows, cover only short-horizon local history, and often emphasize either perceptual details or high-level semantics rather than both. 
Beyond direct encoding of historical observations, some methods exploit abstract or indirect temporal cues. TraceVLA~\cite{zheng2025tracevla} paints historical states as trajectories on the current frame, potentially missing fine-grained details. PTP~\cite{torne2025learning} introduces supervision on past action tokens, providing implicit short-term consistency. 
Language planning methods such as MemER~\cite{sridhar2025memer}, Mem-0~\cite{chen2026rmbench}, and MEM~\cite{torne2026mem} rely on external VLMs and non-end-to-end pipelines, introducing substantial overhead and complicating coordination between high-level planning and low-level VLA. 
In contrast, our method captures long-horizon memory through an end-to-end framework, jointly preserving high-level cognitive semantics and low-level perceptual details, while incorporating imagination for full temporal modeling. 

\subsection{Future Modeling for Robotic Manipulation}
Generative video models~\cite{blattmann2023stable,wan2025wan,ali2025world} provide a promising direction for robotic future modeling, as they encode rich spatio-temporal priors and implicit physical dynamics from large-scale video data. 
One line of work explicitly predicts future visual states to guide action prediction. SuSIE~\cite{black2024zero} generates subgoal images for goal-conditioned control, while UniPi~\cite{du2023learning} formulates policy learning as video generation followed by inverse dynamics. Recent works such as ForeAct~\cite{zhang2026foreact}, VISTA~\cite{long2026scaling}, and $\pi_{0.7}$~\cite{intelligence2026pi} further extend this paradigm to VLA models by jointly modeling future visual prediction and subtask language planning. However, explicit visual prediction can be computationally expensive and may struggle to capture fine-grained physical dynamics. 
Another line of work models visual dynamics and action prediction in a shared latent space. VPP~\cite{hu2024video} and Mimic-Video~\cite{pai2025mimic} adapt video diffusion models to robot data and use their representations as visual policy features. Seer~\cite{tian2025predictive} builds an end-to-end framework that jointly learns future-state prediction and inverse dynamics for action prediction. 
Although these methods efficiently capture fine-grained future dynamics in compact latent spaces, their predictions may contain decision-irrelevant noise, and the lack of explicit memory modeling leads to incomplete temporal representations. 
In contrast, we extract compact future dynamics from partially denoised latents and selectively integrate them under memory guidance, suppressing decision-irrelevant noise while unifying past, present, and future temporal cues. 

\begin{figure*}[t]
    \centering
    \includegraphics[width=1\linewidth]{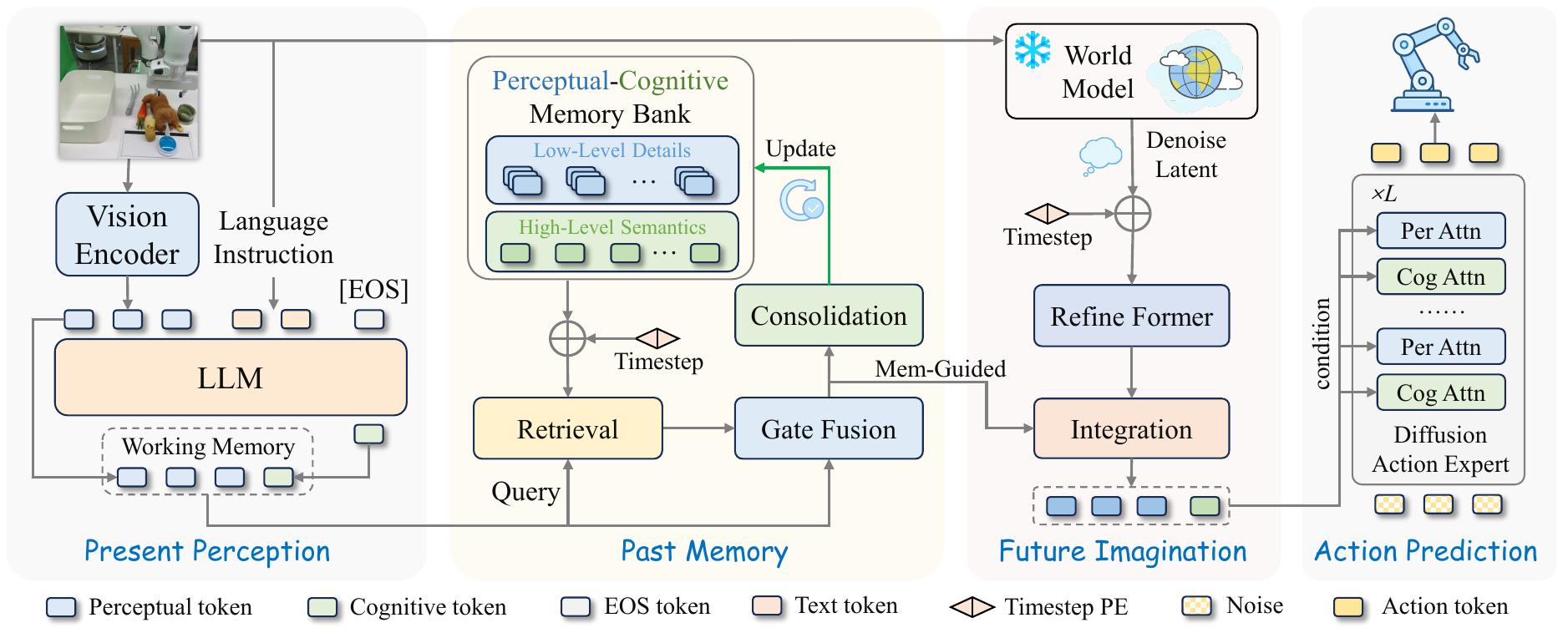}
    \caption{Overall architecture. 
    The current RGB observation and language instruction are encoded by a 7B VLM into perceptual and cognitive tokens, forming working memory.
    The working memory queries a Perceptual-Cognitive Memory Bank (PCMB) to retrieve relevant historical context with high-level semantics and low-level details. 
    The retrieved context is adaptively fused with current tokens, while the PCMB is updated by merging the most similar neighbors.
    A world model imagines future states in a denoising latent space, and the imagined latents are integrated under memory guidance to form full temporal-aware tokens.
    These tokens condition a diffusion action expert to predict temporally coherent action sequences.
    }
    \label{fig:overall}
\end{figure*}

\section{Method}
\label{sec:method}

\subsection{Overview}
\subsubsection{Problem Formulation}
We formulate robotic manipulation with VLA models as a sequential decision-making problem, where visual observations and language instruction are mapped to actions for real-world interaction. Given RGB observations from one or multiple camera views, denoted as \(I = \{I^v\}_{v=1}^{V}\) with \(I^v \in \mathbb{R}^{H \times W \times 3}\), and a language instruction \(L\), a parameterized policy \(\pi\) predicts a sequence of actions:
\begin{equation}
    \mathcal{A} = (a_1, \dots, a_T) = \pi(I, L).
\end{equation}
For single-arm manipulation, each action is defined as
\begin{equation}
    a_t = [\Delta x, \Delta y, \Delta z, \Delta \theta_x, \Delta \theta_y, \Delta \theta_z, g]^\top,
\end{equation}
which contains relative end-effector translation, relative rotation represented by Euler angles, and a binary gripper state \(g \in \{0,1\}\). For dual-arm manipulation, \(a_t\) is defined as the concatenation of the action vectors of both arms.

\subsubsection{Overall Architecture}
MemoryVLA++ is an end-to-end framework for robotic manipulation, as shown in Fig.~\ref{fig:overall}. 
Given the current observation and language instruction, a VLM first encodes them into perceptual and cognitive tokens as working memory. 
To incorporate historical context, we introduce a Perceptual-Cognitive Memory Bank (PCMB) that stores both high-level semantics and fine-grained perceptual details from previous interactions. The working memory queries the PCMB to retrieve decision-relevant historical representations, which are then adaptively fused with the current representations through a gating mechanism. Meanwhile, the PCMB is updated through redundancy-aware consolidation, where temporally adjacent and semantically similar entries are merged when the memory capacity is reached. 
To further model future dynamics, the current observation and language instruction are fed into a pretrained world model, which performs partial denoising to generate multi-scale imagined latents. Guided by the memory-enhanced representations, these imagined latents are adaptively integrated with them to preserve decision-relevant future cues while suppressing irrelevant predictions. 
The resulting full temporal-aware representations are finally used to condition a diffusion action expert to predict a sequence of \(T\) future 7-DoF actions. 

\subsection{Vision-Language-Cognition Module}
The vision-language-cognition module produces perceptual tokens for fine-grained visual details and cognitive token for high-level semantics from visual observations and language instructions. 
We build this module upon a 7B-parameter Prismatic VLM~\cite{karamcheti2024prismatic}, which is further pretrained on the large-scale cross-embodiment real-robot dataset Open-X Embodiment~\cite{o2024open}. 
Given RGB observations from one or multiple camera views, denoted as \(I=\{I^v\}_{v=1}^{V}\), we extract visual features from each view using parallel DINOv2~\cite{oquab2024dinov2} and SigLIP~\cite{zhai2023sigmoid} encoders, and concatenate their outputs as raw visual tokens. 
These tokens are then processed by two parallel branches. 
In the perceptual branch, a SE-bottleneck-based compression module~\cite{hu2018squeeze} reduces the channel dimension of the raw visual tokens and produces perceptual tokens \(p \in \mathbb{R}^{N_p \times d_p}\), which preserve fine-grained visual cues for manipulation. 
In the cognitive branch, the raw visual tokens are projected into the language embedding space and concatenated with the tokenized instruction. 
The resulting multimodal sequence is fed into LLaMA-7B~\cite{touvron2023llama}, and the output at the end-of-sentence (EOS) position is used as the cognitive token \(c \in \mathbb{R}^{1\times d_c}\), capturing compact high-level semantics. 
The perceptual tokens \(p\) and cognitive token \(c\) together form the working memory. 

\subsection{Perceptual-Cognitive Memory Modeling}

\begin{figure*}[t]
    \centering
    \includegraphics[width=0.9\linewidth]{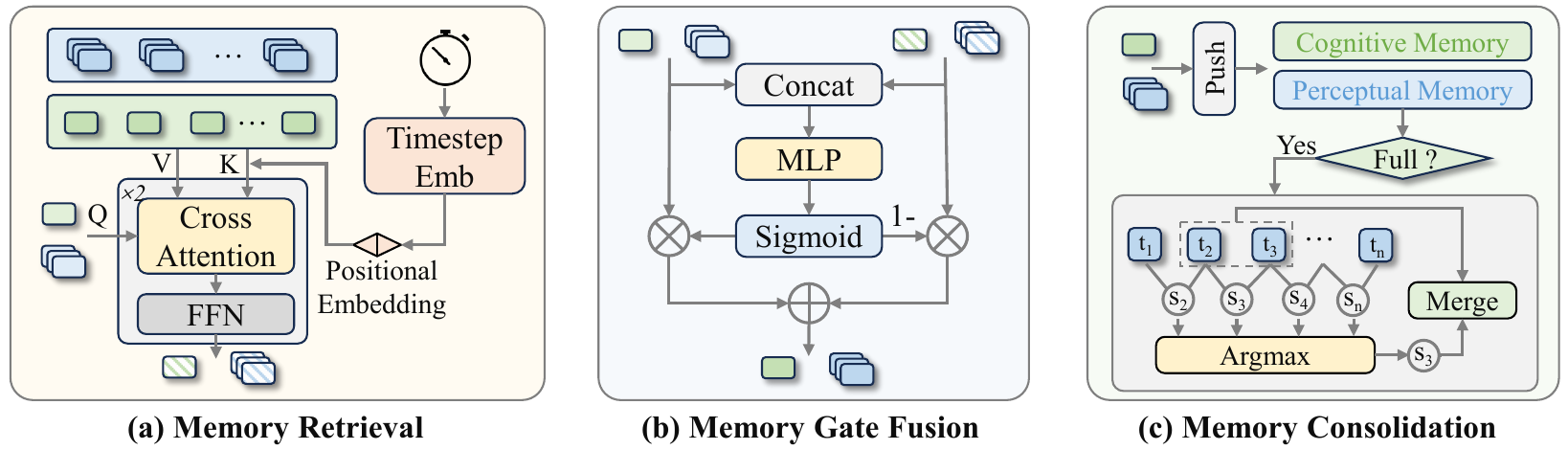}
    \caption{Details of memory module.
    (a) Retrieval: current perceptual and cognitive tokens query the PCMB via cross-attention with timestep PE to fetch relevant historical context. 
    (b) Gate fusion: current tokens and retrieved histories are adaptively fused through a gating mechanism. 
    (c) Consolidation: the current tokens are written back to the PCMB. When the PCMB reaches capacity, the most similar adjacent entries are merged to keep the memory compact. 
    }
    \label{fig:mem_details}
\end{figure*}

The working memory is defined as
\begin{equation}
    M_{\mathrm{wk}} = \{\,p \in \mathbb R^{N_p\times d_p},\;c \in \mathbb R^{1\times d_c}\,\},
\end{equation}
where \(p\) and \(c\) denote the current perceptual tokens and cognitive token, respectively. 
However, this working memory mainly captures the current observation and lacks temporal dependencies. To address this limitation, we introduce the Perceptual-Cognitive Memory Bank (PCMB):
\begin{equation}
    M_{\mathrm{pcmb}} = \{\, m^x \mid x \in \{p,c\}\,\},
\end{equation}
\begin{equation}
    m^x = \{\, m^x_i \in \mathbb R^{N_x \times d_x} \,\}_{i=1}^{L},\quad x \in \{p,c\},
\end{equation}
where \(N_c=1\) for the cognitive stream. 
Each perceptual entry \(m^p_i\) stores fine-grained visual details, while each cognitive entry \(m^c_i\) encodes a compact high-level semantic summary. The PCMB maintains up to \(L\) temporally ordered entries. 

\subsubsection{Memory Retrieval}
At each timestep, the working memory \(M_{\mathrm{wk}}\), consisting of perceptual tokens \(p\) and cognitive token \(c\), acts as the query to retrieve decision-relevant historical information from the PCMB, as illustrated in Fig.~\ref{fig:mem_details}~(a). 
For each stream \(x\in\{p,c\}\), we set the query as
\begin{equation}
    q^p = p,\quad q^c = c.
\end{equation}
Each memory entry is associated with its episode timestep via a sinusoidal timestep embedding \(\mathrm{TE}(\cdot)\), which is projected to the corresponding dimension and added as positional encoding. The keys and values for stream \(x\) are constructed as
\begin{equation}
    K^x = [\,m^x_1 + \mathrm{TE}(t_1);\;\dots;\;m^x_L + \mathrm{TE}(t_L)\,],
\end{equation}
\begin{equation}
    V^x = [\,m^x_1;\;\dots;\;m^x_L\,],
\end{equation}
where the timestep embedding is broadcast when needed. 

The perceptual memory entries are stacked into \(K^p,V^p\in\mathbb{R}^{LN_p\times d_p}\), while the cognitive memory entries are stacked into \(K^c,V^c\in\mathbb{R}^{L\times d_c}\). 
Scaled dot-product attention then retrieves historical representations for each stream:
\begin{equation}
    \hat H^x = \mathrm{softmax}\!\left(
        \frac{q^x (K^x)^\top}{\sqrt{d_x}}
    \right)V^x,\quad x\in\{p,c\}.
\end{equation}
This attention operation is followed by a feed-forward network to form one Transformer layer. After applying two such layers, we obtain the final retrieved embeddings \(H^p\) and \(H^c\). 

\subsubsection{Memory Gate Fusion}
As illustrated in Fig.~\ref{fig:mem_details}~(b), the retrieved embeddings \(H^p\) and \(H^c\) are adaptively fused with the current representations through learned gates. 
For each stream \(x\in\{p,c\}\), the gating vector is computed from the current representation \(x\) and the retrieved embedding \(H^x\):
\begin{equation}
    g^x = \sigma\bigl(\mathrm{MLP}(\mathrm{concat}[x,\,H^x])\bigr).
\end{equation}
The memory-augmented representation is then obtained by
\begin{equation}
    \tilde x = g^x \odot H^x + (1 - g^x) \odot x.
\end{equation}
Here, \(\sigma\) denotes the sigmoid activation and \(\odot\) denotes element-wise multiplication. 
This produces memory-augmented perceptual and cognitive representations \(\tilde p\) and \(\tilde c\). 

\subsubsection{Memory Consolidation} 
After gate fusion, the memory-augmented representations \(\tilde p\) and \(\tilde c\) are passed to downstream modules and simultaneously written into the PCMB. 
When the number of stored entries exceeds the memory capacity \(L\), redundancy-aware consolidation is performed to keep the memory compact, as illustrated in Fig.~\ref{fig:mem_details}~(c). 

For each stream \(x\in\{p,c\}\), cosine similarities are computed between temporally adjacent memory entries. 
The most similar adjacent pair is merged by averaging their representations:
\begin{equation}
\begin{aligned}
    i^*_x &= {\arg\max}_{i=1,\dots,L-1} 
    \cos\!\left(m^x_i,\,m^x_{i+1}\right),\\
    m^x_{i^*_x} &\leftarrow \tfrac{1}{2}
    \left(m^x_{i^*_x}+m^x_{i^*_x+1}\right),
    \quad x \in \{p,c\}.
\end{aligned}
\end{equation}
The merged neighbor \(m^x_{i^*_x+1}\) is then removed from the corresponding stream. 
This mechanism mitigates memory bloat by merging redundant adjacent entries, while maintaining compact memory for long-term modeling. 

\subsection{World Model-Based Imagination Modeling}
\label{sec:imag}

\begin{figure*}[t]
    \centering
    \includegraphics[width=0.9\linewidth]{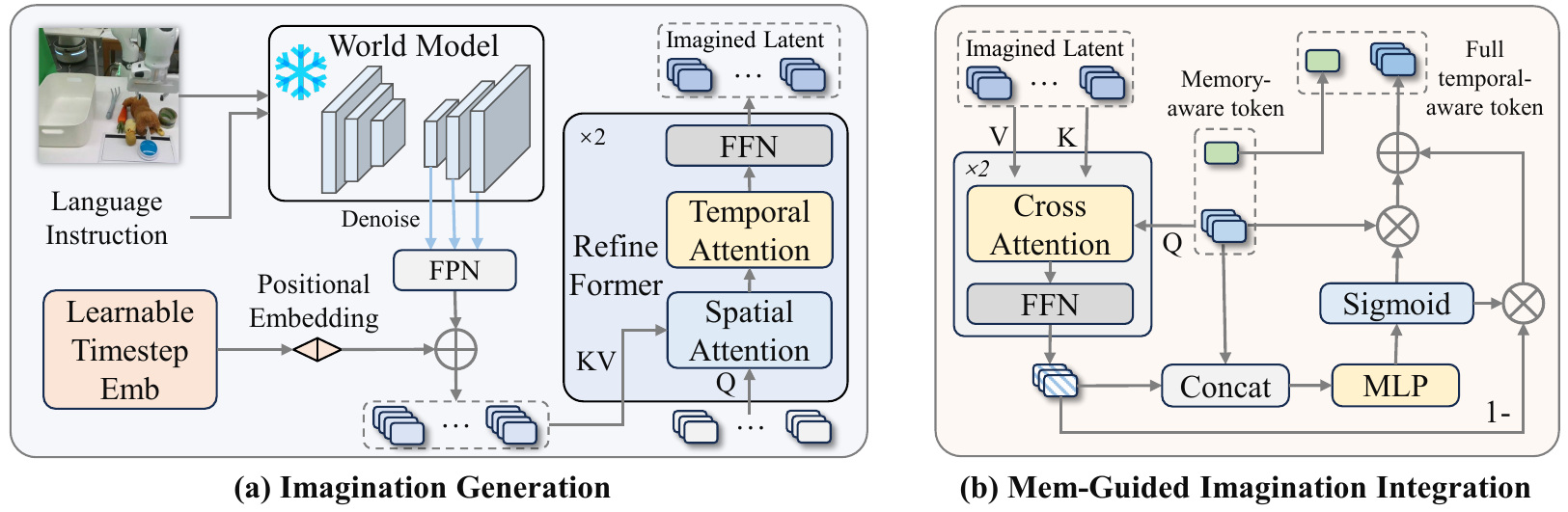}
    \caption{Details of imagination module.
    (a) Imagination generation: conditioned on the current observation and instruction, the world model denoises multi-scale future latent tokens, followed by spatial and temporal attention to capture decision-relevant future state evolution.
    (b) Memory-guided imagination integration: memory-aware tokens attend to imagined latents and adaptively fuse future cues to form full temporal-aware tokens.
    }
    \label{fig:imag_details}
\end{figure*}

Although perceptual-cognitive memory provides historical context from past interactions, robotic manipulation also requires anticipating how the current scene may evolve in the near future. 
Instead of explicitly predicting future RGB frames, which is computationally expensive and often introduces control-irrelevant pixel details, we use a video-generation world model as a latent imagination module. 
The world model extracts compact future cues from partially denoised features, and a memory-guided integration module further suppresses irrelevant imagined content to produce full temporal-aware tokens for action prediction.

\subsubsection{Manipulation-Oriented World Model Adaptation}
\label{sec:imag_adapt}
We instantiate the world model with Stable Video Diffusion (SVD)~\cite{blattmann2023stable}, a 1.5B-parameter video diffusion model pretrained on large-scale Internet videos. 
Following VPP~\cite{hu2024video}, we condition SVD on both the current observation \(I\) and instruction \(L\), where \(L\) is encoded by CLIP and injected into the spatio-temporal UNet via cross-attention. 
Although SVD provides strong visual dynamics priors, we adapt it on manipulation videos to better align with robotic manipulation. 

Given a training sample \((I,L,x_0)\sim\mathcal{D}_{\mathrm{wm}}\), where \(x_0\) denotes the target future video sequence, we sample its noisy version \(x_{\tau}\) at diffusion timestep \(\tau\):
\begin{equation}
    x_{\tau}
    =
    \sqrt{\bar{\alpha}_{\tau}}x_0
    +
    \sqrt{1-\bar{\alpha}_{\tau}}\epsilon,
    \quad
    \epsilon \sim \mathcal{N}(0,1).
\end{equation}
Here, \(\bar{\alpha}_{\tau}\) denotes the cumulative noise schedule. 
The world model \(\mathcal{W}_{\phi}\) is trained to reconstruct the clean future video sequence conditioned on \(I\) and \(L\):
\begin{equation}
    \mathcal{L}_{\mathrm{wm}}
    =
    \mathbb{E}_{(I,L,x_0)\sim\mathcal{D}_{\mathrm{wm}},\,\epsilon,\,\tau}
    \left[
    \left\|
    \mathcal{W}_{\phi}(x_{\tau},\tau,I,L)-x_0
    \right\|_2^2
    \right].
\end{equation}
The adaptation dataset \(\mathcal{D}_{\mathrm{wm}}\) consists of manipulation videos for learning robot-centric priors of future state evolution. 

\subsubsection{Imagination Generation}
\label{sec:imag_gen}
As illustrated in Fig.~\ref{fig:imag_details}~(a), during VLA training, the manipulation-adapted world model is frozen and used only for latent imagination. 
Instead of decoding future RGB frames, we perform partial denoising and extract multi-scale intermediate UNet features \(\{U_s\}_{s=1}^{S}\), where \(U_s\in\mathbb{R}^{K\times C_s\times H_s\times W_s}\). 
Here, \(S\) is the number of feature scales and \(K\) is the number of imagined future steps. 
An FPN~\cite{lin2017feature} is used to aggregate these features into latent tokens:
\begin{equation}
    z
    =
    \mathrm{FPN}(\{U_s\}_{s=1}^{S}),
    \quad
    z\in\mathbb{R}^{K\times N_z\times d_p},
\end{equation}
where \(N_z\) is the number of latent tokens per imagined step and \(d_p\) matches the perceptual-token dimension. 
A learnable temporal embedding \(e_{\mathrm{time}}\in\mathbb{R}^{K\times 1\times d_p}\) is added as
\begin{equation}
    \bar z = z + e_{\mathrm{time}}.
\end{equation}

We then use an imagination former to compress the latent tokens. 
Learnable queries \(q\in\mathbb{R}^{K\times N_q\times d_p}\) attend to the latent tokens through query-based spatial attention:
\begin{equation}
    \hat q_k
    =
    \mathrm{SpatAttn}
    (q_k,\bar z_k,\bar z_k),
    \quad k=1,\dots,K.
\end{equation}
The queries are further processed by temporal attention:
\begin{equation}
    z_{\mathrm{img}}
    =
    \mathrm{FFN}
    \left(
    \mathrm{TempAttn}(\hat q_{1:K})
    \right).
\end{equation}
The resulting \(z_{\mathrm{img}}\in\mathbb{R}^{K\times N_q\times d_p}\) encodes compact future state evolution.

\subsubsection{Imagination Integration}
\label{sec:imag_inte}
Although \(z_{\mathrm{img}}\) captures future dynamics, it may still contain unreliable or decision-irrelevant predictions. 
To suppress such noise, we introduce a memory-guided imagination integration module, as illustrated in Fig.~\ref{fig:imag_details}~(b). 
Specifically, the memory-augmented perceptual tokens \(\tilde p\) query the imagined tokens \(z_{\mathrm{img}}\) through cross-attention, followed by an FFN:
\begin{equation}
    h
    =
    \mathrm{FFN}
    \left(
    \mathrm{CrossAttn}
    (
    \tilde p,
    z_{\mathrm{img}},
    z_{\mathrm{img}}
    )
    \right).
\end{equation}
The representation \(h\) is then adaptively fused with \(\tilde p\):
\begin{equation}
    g
    =
    \sigma
    \bigl(
    \mathrm{MLP}
    (
    \mathrm{concat}[h,\tilde p]
    )
    \bigr),
\end{equation}
\begin{equation}
    \bar p
    =
    g \odot \tilde p
    +
    (1-g)\odot h,
\end{equation}
where \(\sigma\) denotes the sigmoid activation and \(\odot\) denotes element-wise multiplication. 
Finally, \(\bar p\) and \(\tilde c\) are combined as the full temporal-aware tokens:
\begin{equation}
    F_{\mathrm{temp}}=\{\bar p,\tilde c\}.
\end{equation}

\subsection{Full Temporal-Aware Action Expert}
\begin{figure}[t]
    \centering
    \includegraphics[width=0.8\linewidth]{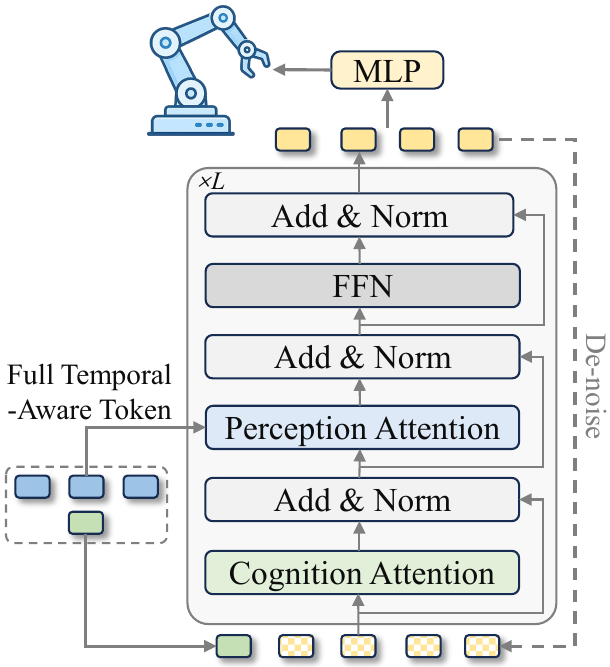}
    \caption{Details of full temporal-aware action expert.
    During diffusion denoising, noisy action tokens are concatenated with cognitive token for cognition attention, while perception attention captures fine-grained details from perceptual tokens for temporally consistent action generation. }
    \label{fig:action_expert}
\end{figure}

As illustrated in Fig.~\ref{fig:action_expert}, the full-temporal-aware tokens \(F_{\mathrm{temp}}=\{\bar p,\tilde c\}\) are used to condition the downstream action expert for action prediction. 
Since robotic control lies in a continuous multimodal action space, we adopt a diffusion-based Transformer (DiT)~\cite{peebles2023scalable} implemented with Denoising Diffusion Implicit Models (DDIM)~\cite{song2020denoising}. 
Starting from noisy action sequences \(\mathcal A_{\tau}\), the model progressively denoises them to predict the target action sequence \(\mathcal A\). 

Specifically, at denoising timestep \(\tau\), the sinusoidal timestep embedding \(\mathrm{TE}(\tau)\) is added to the cognitive token \(\tilde c\), and the resulting representation is concatenated with the noisy action sequence \(\mathcal A_{\tau}\). 
A cognition-attention layer then performs self-attention over the concatenated tokens to provide high-level semantic guidance:
\begin{equation}
    h_c
    =
    \mathrm{CogAttn}
    \bigl(
    [\,\tilde c+\mathrm{TE}(\tau);\;\mathcal A_{\tau}\,]
    \bigr).
\end{equation}
The cognition-aware features further attend to the perceptual tokens \(\bar p\) through a perception-attention layer to inject fine-grained visual details, followed by an FFN:
\begin{equation}
    \hat{\mathcal A}_0
    =
    \mathrm{FFN}
    \left(
    \mathrm{PerAttn}
    (
    h_c,
    \bar p,
    \bar p
    )
    \right).
\end{equation}
The model is trained with mean squared error (MSE) loss between the predicted and target action sequences, and the final denoised representations are projected through an MLP to generate continuous 7-DoF robotic actions. 

\section{Experiments}
\begin{figure*}[t]
    \centering
    \includegraphics[width=1\linewidth]{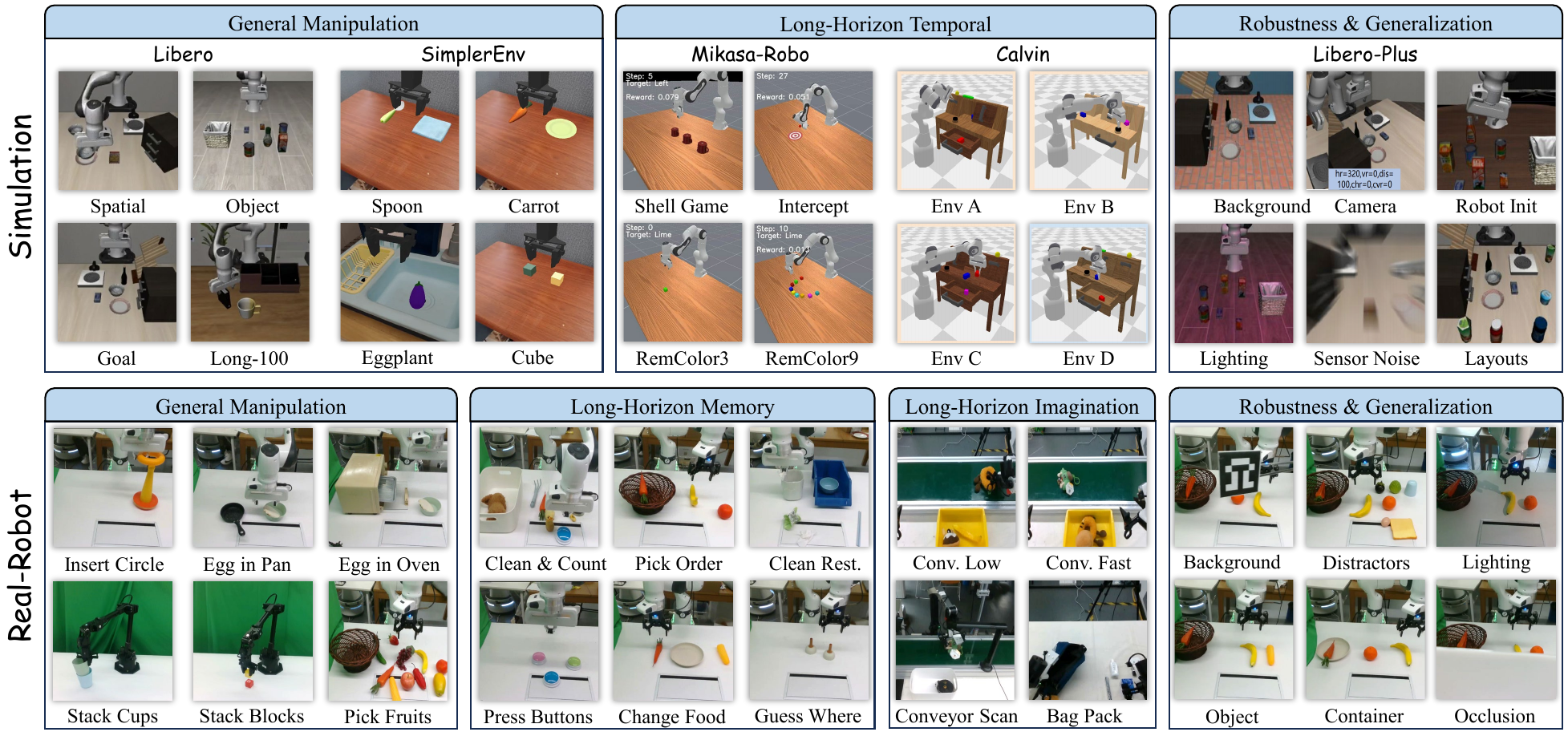}
    \caption{Experimental setup overview. 
    Top: simulation evaluation, covering general manipulation (Libero and SimplerEnv), long-horizon temporal manipulation (Mikasa-Robo and Calvin), and robustness \& generalization evaluation (Libero-Plus). 
    Bottom: real-robot evaluation on general tasks, long-horizon memory-dependent tasks, long-horizon imagination-dependent tasks, and robustness \& generalization settings. 
    Overall, our evaluation spans 3 robots, 5 simulation benchmarks, and 3 categories of real-robot tasks, covering nearly 200 tasks with extensive variations.}
    \label{fig:exp_setup}
\end{figure*}

We conduct extensive experiments to evaluate the effectiveness of MemoryVLA++. 
Secs.~\ref{sec:general}, \ref{sec:temporal}, and~\ref{sec:robust} report simulation results on general manipulation tasks, long-horizon temporal tasks, and robustness settings, respectively. 
Sec.~\ref{sec:real} presents real-robot results on general manipulation tasks, long-horizon memory-dependent tasks, long-horizon imagination-dependent tasks, and robustness settings. 
Finally, Secs.~\ref{sec:ablation},~\ref{sec:analysis}, and~\ref{sec:visual} provide ablation studies, analytical results, and qualitative visualizations.

\subsection{Experimental Setup}
\label{sec:setup}

Fig.~\ref{fig:exp_setup} provides an overview of our simulation and real-world experiments. 
Overall, our experiments span 3 robots, 5 simulation benchmarks, and 3 categories of real-robot tasks, covering nearly 200 tasks with extensive variations. CogACT~\cite{li2024cogact} is used as the primary baseline, and the strongest reported method is used when CogACT results are unavailable.

\begin{figure*}[ht]
    \centering
    \includegraphics[width=1.0\linewidth]{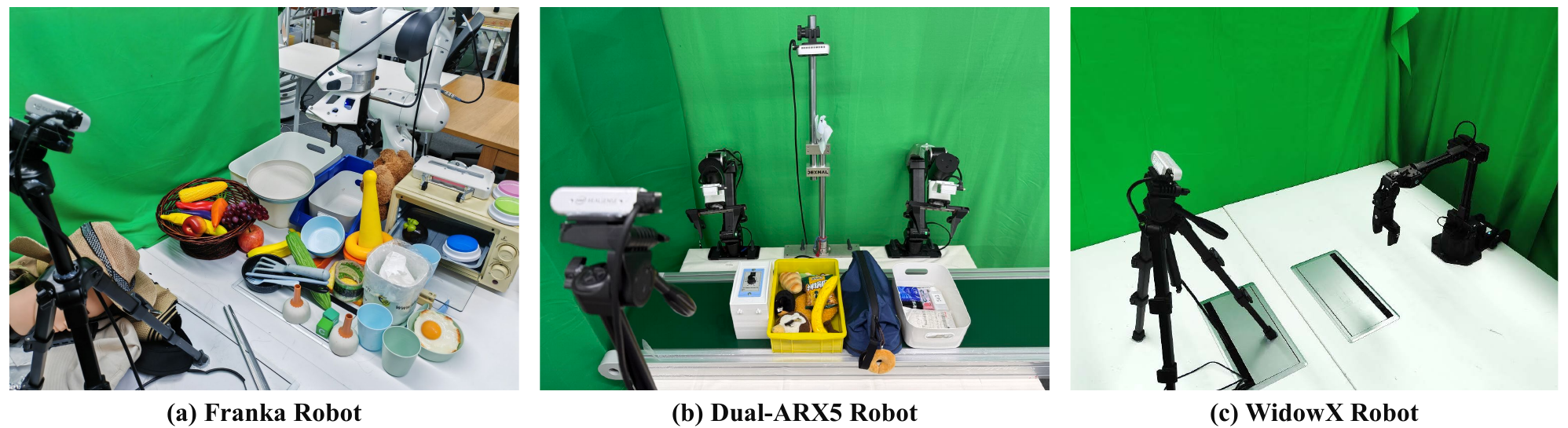}
    \caption{Real-robot platforms used in our experiments: (a) Franka, (b) Dual-ARX5, and (c) WidowX.}
    \label{fig:robot}
\end{figure*}

\subsubsection{Simulation Benchmarks}
We evaluate our method on 5 simulation benchmarks, covering general manipulation, long-horizon temporal manipulation, robustness, and generalization. 

\textbf{Libero}~\cite{liu2023libero} is a manipulation benchmark built with Franka robot. 
It consists of 5 suites: Spatial, Object, Goal, Long-10, and Long-90, totaling 130 tasks. Each task provides 50 demonstrations, covering general and long-horizon tasks. 

\textbf{SimplerEnv}~\cite{li2025evaluating} provides real-to-sim evaluation environments for general robot manipulation. The policy is trained on BridgeData-v2~\cite{walke2023bridgedata}, a large-scale real-robot dataset containing approximately 60,000 teleoperated trajectories collected on WidowX robots across diverse tabletop scenes, and directly evaluated in simulation. 

\textbf{Mikasa-Robo}~\cite{cherepanov2025memory} focuses on temporal robotic manipulation with Franka robot. It includes 5 memory-dependent tasks, each with 250 demonstrations. 

\textbf{Calvin}~\cite{mees2022calvin} evaluates long-horizon language-conditioned manipulation, requiring the robot to complete 5 consecutive instructions by composing multiple skills. We follow the ABC$\rightarrow$D protocol, training on environments A, B, and C and testing exclusively on environment D. 

\textbf{Libero-Plus}~\cite{fei2025libero} extends Libero with additional variations for robustness and generalization evaluation, including camera view, robot initialization, language instruction, lighting, background, noise, and layout variations. 

\subsubsection{Real-Robot Setup} 
We further evaluate our method on 3 real-robot platforms, including Franka, WidowX, and Dual-ARX5, as shown in Fig.~\ref{fig:robot}. 
The real-robot experiments cover 3 categories: general manipulation, long-horizon memory-dependent manipulation, and long-horizon imagination-dependent manipulation. 
Franka and WidowX use a fixed three-view RGB setup with Intel RealSense D435 cameras, while Dual-ARX5 uses fixed RGB cameras together with an additional wrist-mounted RGB camera. 
RGB observations are captured at $640\times480$ resolution and 30\,fps. 
Demonstrations are collected through teleoperation, and the robot system is integrated with ROS. 
After collection, frames are downsampled to $224\times224$ and temporally subsampled by retaining a frame whenever the end-effector translation since the last kept frame exceeds $0.01$\,m or the orientation change exceeds $0.4$\,rad. 
The processed episodes are converted into the RLDS format for downstream training.

\subsubsection{Implementation Details} 
We train on 8 NVIDIA A100 or H20 GPUs with PyTorch FSDP, using 26-32 samples per GPU for a global batch of 208-256 and a learning rate of $2\times10^{-5}$. 
The LLM has 7B parameters, and the diffusion action expert contains approximately 300M parameters. During inference, we use DDIM~\cite{song2020denoising} with 10 sampling steps and classifier-free guidance (CFG)~\cite{ho2022classifier} with a guidance scale of 1.5.

\begin{table}[t]
  \centering
  \caption{General robotic manipulation performance on Libero~\cite{liu2023libero}. Success rates (\%) are reported across 5 suites. For methods without Long-90 results, the average is computed over the first 4 suites. \textbf{Bold} denotes the best result, and \underline{underline} denotes the baseline.}
  \label{tab:libero}
  \resizebox{1.0\linewidth}{!}{%
    \begin{tabular}{lcccccc}
      \toprule
      Method & Spatial & Object & Goal & Long-10 & Long-90 & \makecell[c]{Avg.\\Succ.} \\
      \midrule
      DP~\scriptsize\cite{chi2023diffusion} & 78.3 & 92.5 & 68.3 & 50.5 & - & 72.4 \\
      Octo~\scriptsize\cite{team2023octo} & 78.9 & 85.7 & 84.6 & 51.1 & - & 75.1 \\
      MDT~\scriptsize\cite{reuss2024multimodal} & 78.5 & 87.5 & 73.5 & 64.8 & - & 76.1 \\
      UniACT~\scriptsize\cite{zheng2025universal} & 77.0 & 87.0 & 77.0 & 70.0 & 73.0 & 76.8 \\
      MaIL~\scriptsize\cite{jia2024mail} & 74.3 & 90.1 & 81.8 & 78.6 & - & 83.5 \\
      SpatialVLA~\scriptsize\cite{qu2025spatialvla} & 88.2 & 89.9 & 78.6 & 55.5 & 46.2 & 71.7 \\
      TraceVLA~\scriptsize\cite{zheng2025tracevla} & 84.6 & 85.2 & 75.1 & 54.1 & - & 74.8 \\
      OpenVLA~\scriptsize\cite{kim2025openvla} & 84.7 & 88.4 & 79.2 & 53.7 & 73.5 & 75.9 \\
      WorldVLA~\scriptsize\cite{cen2025worldvla} & 87.6 & 96.2 & 83.4 & 60.0 & - & 81.8 \\
      $\pi_0$-FAST~\scriptsize\cite{pertsch2025fast} & 96.4 & 96.8 & 88.6 & 60.2 & 83.1 & 85.0 \\
      TriVLA~\scriptsize\cite{liu2025trivla} & 91.2 & 93.8 & 89.8 & 73.2 & - & 87.0 \\
      SmolVLA~\scriptsize\cite{shukor2025smolvla} & 93.0 & 94.0 & 91.0 & 77.0 & - & 88.8 \\
      4D-VLA~\scriptsize\cite{zhang20264d} & 93.8 & 92.8 & 95.6 & 86.5 & - & 92.2 \\
      DreamVLA~\scriptsize\cite{zhang2025dreamvla} & 97.5 & 94.0 & 89.5 & 89.5 & - & 92.6 \\
      CogACT~\scriptsize\cite{li2024cogact} & 97.2 & 98.0 & 90.2 & 88.8 & 92.1 & \underline{93.2} \\
      UniVLA~\scriptsize\cite{wang2025unified} & 96.4 & 98.0 & 90.8 & 89.6 & - & 93.7 \\
      GR00T-N1.5~\scriptsize\cite{bjorck2025gr00t} & 94.4 & 97.6 & 93.0 & 90.6 & - & 93.9 \\
      $\pi_0$~\scriptsize\cite{black2025pi_0} & 96.8 & 98.8 & 95.8 & 85.2 & - & 94.2 \\

      \midrule
      \rowcolor{gray!15} MemoryVLA
        & 98.4 & 98.4 & 96.4 & 93.4 & 95.6
        & \textbf{96.5} \;\gdelta{(+3.3)} \\
      \rowcolor{gray!15} MemoryVLA++
        & 99.8 & 100.0 & 98.2 & 96.0 & 97.8
        & \textbf{98.4} \;\gdelta{(+5.2)} \\
      \bottomrule
    \end{tabular}%
  }
\end{table}

\begin{table}[t]
  \centering
  \caption{General robotic manipulation performance on SimplerEnv~\cite{li2025evaluating}. $\pi_0$ is reproduced using \href{https://github.com/allenzren/open-pi-zero}{open-pi-zero}. 
  CogACT is evaluated with the official checkpoint. }
  \label{tab:simpler}
  
  \resizebox{\columnwidth}{!}{%
    \begin{tabular}{lccccc}
      \toprule
      Method
        & \makecell[c]{Spoon\\on Towel}
        & \makecell[c]{Carrot\\on Plate}
        & \makecell[c]{Stack\\Cube}
        & \makecell[c]{Eggplant\\in Basket}
        & \makecell[c]{Avg.\\Succ.} \\
      \midrule
      RT-1-X~\scriptsize\cite{o2024open}
        & 0.0 & 4.2 & 0.0 & 0.0 & 1.1 \\
      OpenVLA~\scriptsize\cite{kim2025openvla}
        & 4.2 & 0.0 & 0.0 & 12.5 & 4.2 \\
      Octo-Base~\scriptsize\cite{team2023octo}
        & 15.8 & 12.5 & 0.0 & 41.7 & 17.5 \\
      TraceVLA~\scriptsize\cite{zheng2025tracevla}
        & 12.5 & 16.6 & 16.6 & 65.0 & 27.7 \\
      OpenVLA-OFT~\cite{kim2025fine}
        & 12.5 & 4.2 & 4.2 & 100.0 & 30.2 \\
      RoboVLMs~\scriptsize\cite{li2026matters}
        & 45.8 & 20.8 & 4.2 & 79.2 & 37.5 \\
      SpatialVLA~\scriptsize\cite{qu2025spatialvla}
        & 16.7 & 25.0 & 29.2 & 100.0 & 42.7 \\
      Magma~\scriptsize\cite{yang2025magma}
        & 37.5 & 29.2 & 20.8 & 91.7 & 44.8 \\
      $\pi_0$-FAST~\cite{pertsch2025fast}
        & 62.5 & 29.2 & 20.8 & 83.3 & 49.0 \\
      DreamVLA~\scriptsize\cite{zhang2025dreamvla}
        & 45.8 & 45.8 & 25.0 & 87.5 & 51.0 \\
      VideoVLA~\scriptsize\cite{shen2025videovla}
        & 75.0 & 20.8 & 45.8 & 70.8 & 53.1 \\
      Mimic-video~\scriptsize\cite{pai2025mimic}
        & 41.7 & 54.2 & 29.2 & 100.0 & 56.3 \\
      CogACT~\scriptsize\cite{li2024cogact}
        & 58.3 & 45.8 & 29.2 & 95.8 & \underline{57.3} \\
      CronusVLA~\scriptsize\cite{li2025cronusvla}
        & 66.7 & 54.2 & 20.8 & 100.0 & 60.4 \\
      GR00T-N1.5~\cite{bjorck2025gr00t}
        & 75.3 & 54.3 & 57.0 & 61.3 & 61.9 \\
      $\pi_0$~\scriptsize\cite{black2025pi_0} 
        & 84.6 & 55.8 & 47.9 & 85.4 & 68.4 \\
      \midrule
      \rowcolor{gray!15} MemoryVLA
        & 75.0 & 75.0 & 37.5 & 100.0
        & \textbf{71.9}\,\gdelta{(+14.6)} \\
      \rowcolor{gray!15} MemoryVLA++
        & 83.3 & 66.7 & 45.8 & 100.0
        & \textbf{73.9}\,\gdelta{(+16.6)} \\
      \bottomrule
    \end{tabular}%
  }
\end{table}

\begin{table}[t]
  \centering
  \caption{Temporal robotic manipulation performance on Mikasa-Robo~\cite{cherepanov2025memory}. Numbers in parentheses indicate the number of input frames per forward pass. SGT, IM, RC3, RC5, and RC9 denote ShellGameTouch, InterceptMedium, RememberColor3, RememberColor5, and RememberColor9, respectively.}
  \label{tab:mikasa}
  \resizebox{\linewidth}{!}{%
    \begin{tabular}{lcccccc}
      \toprule
      Method & SGT & IM & RC3 & RC5 & RC9 & \makecell[c]{Avg.\\Succ.} \\
      \midrule
      \rowcolor{black!4}
      \multicolumn{7}{c}{\textit{Multi-frame policies}} \\
      DP (4)~\scriptsize\cite{chi2023diffusion}
      & 23 & - & 3 & - & - & - \\
      DP (8)~\scriptsize\cite{chi2023diffusion}
      & 18 & - & 7 & - & - & - \\
      PTP (4)~\scriptsize\cite{torne2025learning}
      & 22 & - & 3 & - & - & - \\
      PTP (8)~\scriptsize\cite{torne2025learning}
      & 26 & - & 4 & - & - & - \\
      MaIL (4)~\scriptsize\cite{jia2024mail}
      & 28 & - & 10 & - & - & - \\
      MaIL (8)~\scriptsize\cite{jia2024mail}
      & 27 & - & 11 & - & - & - \\
      Octo (10)~\scriptsize\cite{team2023octo}
      & 46 & 39 & 45 & 17 & 11 & 31.6 \\

      \midrule
      \rowcolor{black!4}
      \multicolumn{7}{c}{\textit{Single-frame VLAs}} \\
      CronusVLA (1)~\scriptsize\cite{li2025cronusvla}
      & 32 & 5 & 31 & 13 & 9 & 18.0 \\
      SpatialVLA (1)~\scriptsize\cite{qu2025spatialvla}
      & 23 & 27 & 27 & 17 & 11 & 21.0 \\
      OpenVLA-OFT (1)~\scriptsize\cite{kim2025fine}
      & 47 & 14 & 59 & 16 & 6 & 28.4 \\
      $\pi_0$ (1)~\scriptsize\cite{black2025pi_0}
      & 33 & 42 & 35 & 22 & 15 & \underline{29.4} \\

      \midrule
      \rowcolor{gray!15}
      MemoryVLA (1)
      & 88 & 24 & 44 & 30 & 20 & \textbf{41.2} \;\gdelta{(+11.8)} \\
      \rowcolor{gray!15}
      MemoryVLA++ (1)
      & 97 & 40 & 50 & 19 & 16 & \textbf{44.4} \;\gdelta{(+15.0)} \\
      \bottomrule
    \end{tabular}%
  }
\end{table}

\begin{table}[t]
  \centering
  \caption{Long-horizon robotic manipulation performance on Calvin~\cite{mees2022calvin} ABC$\rightarrow$D setting. We report success rates over 1000 rollouts and the average number of completed tasks. CogACT results follow~\cite{xie2025dexbotic}. }
  \label{tab:calvin}
  \resizebox{\linewidth}{!}{%
  \begin{tabular}{lcccccc}
    \toprule
    \multirow{2}{*}[-0.7ex]{Method}
      & \multicolumn{6}{@{\hspace{4pt}\vrule\hspace{4pt}}c}{Task Success Rate} \\
    \cmidrule(lr){2-7}
      & \multicolumn{1}{@{\hspace{4pt}\vrule\hspace{4pt}}c}{1}
      & 2 & 3 & 4 & 5 & Avg. Len. $\uparrow$ \\
    \midrule
    \rowcolor{black!4}
    \multicolumn{7}{c}{\textit{ABC$\rightarrow$D Zero-Shot Setting}} \\
    DP~\scriptsize\cite{chi2023diffusion}
      & 40.2 & 12.3 & 2.6 & 0.8 & 0.0 & 0.56 \\
    RT-1~\scriptsize\cite{brohan2023rt}
      & 53.3 & 22.2 & 9.4 & 3.8 & 1.3 & 0.90 \\
    UniPi~\scriptsize\cite{du2023learning}
      & 56.0 & 16.0 & 8.0 & 8.0 & 4.0 & 0.92 \\
    Roboflamingo~\scriptsize\cite{li2024vision}
      & 82.4 & 61.9 & 46.6 & 33.1 & 23.5 & 2.47 \\
    DeerVLA~\scriptsize\cite{yue2024deer}
      & 89.7 & 70.5 & 51.8 & 44.2 & 35.3 & 2.92 \\
    GR-1~\scriptsize\cite{wu2023unleashing}
      & 85.4 & 71.2 & 59.6 & 49.7 & 40.1 & 3.06 \\
    Moto~\scriptsize\cite{chen2025moto}
      & 89.7 & 72.9 & 60.1 & 48.4 & 38.6 & 3.10 \\
    OpenVLA~\scriptsize\cite{kim2025openvla}
      & 91.3 & 77.8 & 62.0 & 52.1 & 43.5 & 3.27 \\
    CogACT~\scriptsize\cite{li2024cogact}
      & 83.8 & 72.9 & 64.0 & 55.9 & 48.0 & \underline{3.25} \\
    OpenVLA-OFT~\scriptsize\cite{kim2025fine}
      & 89.1 & 79.4 & 67.4 & 59.8 & 51.5 & 3.47 \\
    CLOVER~\scriptsize\cite{bu2024closed}
      & 96.0 & 83.5 & 70.8 & 57.5 & 45.4 & 3.53 \\
    RoboDual~\scriptsize\cite{bu2024towards}
      & 94.4 & 82.7 & 72.1 & 62.4 & 54.4 & 3.66 \\
    UniVLA~\scriptsize\cite{bu2025univla}
      & 95.5 & 85.8 & 75.4 & 66.9 & 56.5 & 3.80 \\
    $\pi_0$~\scriptsize\cite{black2025pi_0}
      & 93.8 & 85.0 & 76.7 & 68.1 & 59.9 & 3.92 \\
    \midrule
    \rowcolor{gray!15} MemoryVLA 
    & 94.8 & 87.4 & 81.4 & 75.9 & 69.4 & \textbf{4.09} \;\gdelta{(+0.84)} \\
    \rowcolor{gray!15} MemoryVLA++
    & 95.6 & 90.2 & 85.7 & 81.7 & 76.1 & \textbf{4.29} \;\gdelta{(+1.04)} \\
    \bottomrule
  \end{tabular}
  }
\end{table}

\subsection{Simulation Evaluation on General Robotic Manipulation}
\label{sec:general}

\subsubsection{Training and Evaluation Details}
For Libero~\cite{liu2023libero}, we evaluate MemoryVLA++ across 5 suites: Spatial, Object, Goal, Long-10, and Long-90. The first 4 suites contain 10 tasks each, while Long-90 contains 90 tasks. Following OpenVLA~\cite{kim2025openvla}, we use 50 demonstrations per task. We jointly train MemoryVLA++ on the first 4 suites for 60k steps and train Long-90 separately for 30k steps. Validation is performed every 2k steps, and results are reported at the best validation checkpoint. Each task is evaluated over 50 trials. 
For SimplerEnv~\cite{li2025evaluating}, we train on the BridgeData-v2 dataset~\cite{walke2023bridgedata} for 50k steps, with validation performed every 2.5k steps. 
Results are reported at the best validation checkpoint, and each task is evaluated over 24 trials. 
For both benchmarks, the corresponding world model is trained for 40k steps.

\subsubsection{Results on Libero}

As shown in Tab.~\ref{tab:libero}, MemoryVLA++ achieves the best overall performance across all suites, with an average success rate of 98.4\%. 
It outperforms the CogACT baseline by +5.2 points, with a particularly large gain of +7.2 points on Long-10. 
Across individual suites, MemoryVLA++ obtains 99.8\% on Spatial, 100.0\% on Object, 98.2\% on Goal, 96.0\% on Long-10, and 97.8\% on Long-90. 
The gains are consistent across both general and long-horizon suites. 

\subsubsection{Results on SimplerEnv}

As shown in Tab.~\ref{tab:simpler}, MemoryVLA++ achieves an average success rate of 73.9\% on SimplerEnv, outperforming CogACT by +16.6 points. 
It obtains 83.3\%, 66.7\%, 45.8\%, and 100.0\% on \textit{Spoon on Towel}, \textit{Carrot on Plate}, \textit{Stack Cube}, and \textit{Eggplant in Basket}, respectively. 

\subsection{Simulation Evaluation on Long-Horizon Temporal Tasks}
\label{sec:temporal}

\subsubsection{Training and Evaluation Details}
We evaluate long-horizon temporal tasks on Mikasa-Robo~\cite{cherepanov2025memory} and Calvin~\cite{mees2022calvin}. 
For Mikasa-Robo, we follow the standard protocol with 250 demonstrations per task, $128\times128$ image observations, end-effector control, and 100 evaluation episodes per task. 
All 5 tasks are jointly trained for 20k steps, validated every 1k steps, and reported at the best validation checkpoint. The world model for Mikasa-Robo is trained for 20k steps. 
As shown in Tab.~\ref{tab:mikasa}, we compare with both multi-frame policies and single-frame VLAs.
For Calvin, we use the ABC$\rightarrow$D zero-shot setting, where policies are trained on environments A, B, and C, and evaluated on the unseen environment D. 
Models are trained for 60k steps, evaluated every 2k steps, and reported at the best evaluation checkpoint. For the world model, we directly follow VPP~\cite{hu2024video}.
We report success rates over 1000 rollouts and the average number of completed tasks. 

\subsubsection{Results on Mikasa-Robo}

As shown in Tab.~\ref{tab:mikasa}, MemoryVLA++ achieves the best average success rate of 44.4\%, outperforming the previous best single-frame VLA by +15.0 points. 
Notably, MemoryVLA++ improves ShellGameTouch by +50.0 points over the previous best single-frame VLA, demonstrating the effectiveness of full temporal modeling.

\subsubsection{Results on Calvin}

As shown in Tab.~\ref{tab:calvin}, MemoryVLA++ achieves the best average completed task length of 4.29, improving over CogACT baseline by +1.04. 
It also achieves higher success rates across all 5 task steps, with larger gains at longer horizons, showing improved temporal consistency in long-horizon task execution. 

\subsection{Simulation Evaluation on Robustness and Generalization}
\label{sec:robust}

\subsubsection{Training and Evaluation Details}
We evaluate robustness and generalization on Libero-Plus~\cite{fei2025libero}. 
Libero-Plus introduces 7 out-of-distribution variations, covering camera, robot, language, lighting, background, noise, and layout. 
In the zero-shot setting, models are trained on the 4 standard Libero suites, including Spatial, Object, Goal, and Long-10. We jointly train on these 4 suites for 60k steps and evaluate on Libero-Plus without using its training data. 
In the supervised fine-tuning setting, Libero-Plus data are included in mixed training, and models are trained for the same 60k steps. 
The world model is trained for 40k steps.
For both settings, evaluation is performed every 10k steps, and we report the best evaluation result. 

\subsubsection{Results on Libero-Plus}

\begin{table*}[ht]
  \centering
  \caption{Robustness \& generalization performance on Libero-Plus~\cite{fei2025libero}. OpenVLA-OFT and our method are both jointly trained on all suites and use wrist image. ``Supervised Fine-Tuning'' includes Libero-Plus data in mixed training. Avg. Succ. is computed over all trials across settings, rather than by averaging setting-level success rates. }
  \label{tab:libero_plus}
  \resizebox{0.75\linewidth}{!}{%
    \begin{tabular}{lcccccccc}
      \toprule
      Method & Camera & Robot & Language & Light & Background & Noise & Layout & Avg. Succ. \\
      \midrule
      \rowcolor{black!4}
      \multicolumn{9}{c}{\textit{Zero-Shot Setting}} \\
      OpenVLA~\scriptsize\cite{kim2025openvla}
      & 0.8  & 3.5  & 23.0 & 8.1  & 34.8 & 15.2 & 28.5 & 15.6 \\
      WorldVLA~\scriptsize\cite{cen2025worldvla}
      & 0.1  & 27.9 & 41.6 & 43.7 & 17.1 & 10.9 & 38.0 & 25.0 \\
      DP~\scriptsize\cite{chi2023diffusion}
      & 1.6 & 32.3 & 77.0 & 25.0 & 19.8 & 20.3 & 42.2 & 31.7 \\
      NORA~\scriptsize\cite{hung2025nora}
      & 2.2  & 37.0 & 65.1 & 45.7 & 58.6 & 12.8 & 62.1 & 39.0 \\
      UniVLA~\scriptsize\cite{bu2025univla}
      & 1.8  & 46.2 & 69.6 & 69.0 & 81.0 & 21.2 & 31.9 & 43.9 \\
      Fast-WAM~\scriptsize\cite{yuan2026fast}
      & 16.4 & 44.5 & 68.9 & 78.2 & 53.7 & 37.7 & 60.7 & 51.5 \\
      $\pi_0$~\scriptsize\cite{black2025pi_0}
      & 13.8 & 6.0  & 58.8 & 85.0 & 81.4 & 79.0 & 68.9 & 53.6 \\
      $\pi_0$-Fast~\scriptsize\cite{pertsch2025fast}
      & 65.1 & 21.6 & 61.0 & 73.2 & 73.2 & 74.4 & 68.8 & 61.6 \\
      OpenVLA-OFT~\scriptsize\cite{kim2025fine}
      & 55.6 & 21.7 & 81.0 & 92.7 & 91.0 & 78.6 & 68.7 & \underline{67.9} \\
      RIPT-VLA~\scriptsize\cite{tan2025interactive}
      & 55.2 & 31.2 & 77.6 & 88.4 & 91.6 & 73.5 & 74.2 & 68.4 \\
      \midrule
      \rowcolor{gray!15} MemoryVLA
      & 42.7 & 44.9 & 84.4 & 92.8 & 95.0 & 62.1 & 84.7 & \textbf{70.2} \\
      \rowcolor{gray!15} MemoryVLA++
      & 36.4 & 68.9 & 88.7 & 93.8 & 90.6 & 63.5 & 83.8 & \textbf{73.1} \\

      \midrule
      \rowcolor{black!4}
      \multicolumn{9}{c}{\textit{Supervised Fine-Tuning Setting}} \\
      $\pi_0$~\scriptsize\cite{black2025pi_0}
      & 79.6 & 21.1 & 72.5 & 84.7 & 86.2 & 68.3 & 69.4 & 67.4 \\
      OpenVLA-OFT~\scriptsize\cite{kim2025fine}
      & 92.8 & 30.3 & 85.8 & 94.9 & 93.9 & 89.3 & 77.6 & \underline{79.6} \\
      \midrule
      \rowcolor{gray!15} MemoryVLA
      & 91.4 & 48.6 & 79.4 & 95.2 & 95.3 & 94.0 & 75.7 & \textbf{81.9} \\
      \rowcolor{gray!15} MemoryVLA++
      & 96.8 & 49.7 & 71.0 & 96.6 & 97.0 & 96.0 & 78.6 & \textbf{82.7} \\
      \bottomrule
    \end{tabular}%
  }
\end{table*}

As shown in Tab.~\ref{tab:libero_plus}, MemoryVLA++ achieves an average success rate of 73.1\% in the zero-shot setting, outperforming OpenVLA-OFT by +5.2 points. 
In the supervised fine-tuning setting, MemoryVLA++ further improves the average success rate to 82.7\%, exceeding OpenVLA-OFT by +3.1 points. 
These results indicate that MemoryVLA++ maintains robustness and generalization under diverse distribution shifts. 

\subsection{Real-Robot Evaluation}
\label{sec:real}

\subsubsection{Real-Robot Tasks}
We evaluate real-robot performance across 3 task categories, as shown in Fig.~\ref{fig:vis_real}.
The general tasks evaluate short-horizon manipulation skills, including \textit{Insert Circle}, \textit{Egg in Pan}, \textit{Egg in Oven}, \textit{Stack Cups}, \textit{Stack Blocks}, and \textit{Pick Diverse Fruits}. 
The long-horizon memory-dependent tasks require the policy to use historical information over multiple sub-goals, including \textit{Push Buttons}, \textit{Change Food}, \textit{Guess Where}, \textit{Clean Table \& Count}, \textit{Pick Place Order}, and \textit{Clean Restaurant Table}.
The long-horizon imagination-dependent tasks require anticipating future state evolution, including \textit{Conveyor Pick-Low}, \textit{Conveyor Pick-Mid}, \textit{Conveyor Pick-High}, \textit{Conveyor Scan-Pick}, and \textit{Bag Pack \& Zip}.

\subsubsection{Training and Evaluation Details}
All tasks are evaluated from randomized initial states. 
For general tasks, each task uses 50-150 demonstrations. \textit{Pick Diverse Fruits} contains 5 variants with 5 trials per variant, resulting in 25 trials in total, while other general tasks use 15 trials.
For long-horizon memory-dependent and imagination-dependent tasks, each task uses 200-300 demonstrations. \textit{Push Buttons} contains 3 variants with 5 trials per variant, resulting in 15 trials in total, while other long-horizon tasks use 10 trials.
We use step-wise scoring for these long-horizon tasks to better reflect progress over multiple sub-goals. 
Training runs for approximately 5k-30k steps depending on the task. 
For MemoryVLA++, the world model is trained for 20k steps. MemoryVLA++ is evaluated on long-horizon imagination-dependent tasks, where future imagination is explicitly required.

\begin{table}[t]
  \centering
  \caption{Real-robot performance across three task categories. Success scores (\%) are reported. MemoryVLA++ is evaluated only on imagination-dependent tasks, where future imagination is explicitly required.}
  \label{tab:real}
  \scriptsize
  \setlength{\tabcolsep}{2.5pt}
  \renewcommand{\arraystretch}{1.05}

  \subfloat[General manipulation tasks.]{
  \resizebox{\columnwidth}{!}{%
    \begin{tabular}{lccccccc}
      \toprule
      Method
        & \makecell[c]{Insert\\Circle}
        & \makecell[c]{Egg in\\Pan}
        & \makecell[c]{Egg in\\Oven}
        & \makecell[c]{Stack\\Cups}
        & \makecell[c]{Stack\\Blocks}
        & \makecell[c]{Pick Diverse\\Fruits}
        & \makecell[c]{Avg.\\Succ.} \\
      \midrule
      OpenVLA~\scriptsize\cite{kim2025openvla}
        & 47 & 27 & 53 & 40 & 13 & 4 & 31 \\
      $\pi_0$~\scriptsize\cite{black2025pi_0}
        & 67 & 73 & 73 & 87 & 53 & 80 & 72 \\
      CogACT~\scriptsize\cite{li2024cogact}
        & 80 & 67 & 60 & 93 & 80 & 76 & \underline{76} \\
      \midrule
      \rowcolor{gray!15} MemoryVLA
        & 87 & 80 & 80 & 93 & 87 & 84 & \textbf{85} \;\gdelta{(+9)} \\
      \bottomrule
    \end{tabular}%
  }}

  \subfloat[Long-horizon memory-dependent tasks.]{
  \resizebox{\columnwidth}{!}{%
    \begin{tabular}{lccccccc}
      \toprule
      Method
        & \makecell[c]{Push\\Buttons}
        & \makecell[c]{Change\\Food}
        & \makecell[c]{Guess\\Where}
        & \makecell[c]{Clean Table\\\& Count}
        & \makecell[c]{Pick Place\\Order}
        & \makecell[c]{Clean Rest.\\Table}
        & \makecell[c]{Avg.\\Succ.} \\
      \midrule
      OpenVLA~\scriptsize\cite{kim2025openvla}
        & 6 & 3 & 0 & 15 & 27 & 0 & 9 \\
      $\pi_0$~\scriptsize\cite{black2025pi_0}
        & 25 & 42 & 24 & 61 & 82 & 80 & 52 \\
      CogACT~\scriptsize\cite{li2024cogact}
        & 15 & 47 & 40 & 67 & 90 & 84 & \underline{57} \\
      \midrule
      \rowcolor{gray!15} MemoryVLA
        & 58 & 85 & 72 & 84 & 100 & 96 & \textbf{83} \;\gdelta{(+26)} \\
      \bottomrule
    \end{tabular}%
  }}

  \subfloat[Long-horizon imagination-dependent tasks.]{
  \resizebox{\columnwidth}{!}{%
    \begin{tabular}{lcccccc}
      \toprule
      Method
        & \makecell[c]{Conveyor\\Pick-Low}
        & \makecell[c]{Conveyor\\Pick-Mid}
        & \makecell[c]{Conveyor\\Pick-High}
        & \makecell[c]{Conveyor\\Scan-Pick}
        & \makecell[c]{Bag Pack\\\& Zip}
        & \makecell[c]{Avg.\\Succ.} \\
      \midrule
      $\pi_0$~\scriptsize\cite{black2025pi_0}
        & 56 & 52 & 44 & 50 & 41 & 49 \\
      CogACT~\scriptsize\cite{li2024cogact}
        & 61 & 55 & 48 & 42 & 37 & \underline{49} \\
      \midrule
      \rowcolor{gray!15} MemoryVLA
        & 75 & 68 & 58 & 63 & 60 & \textbf{65} \;\gdelta{(+16)} \\
      \rowcolor{gray!15} MemoryVLA++
        & 84 & 80 & 72 & 75 & 73 & \textbf{77} \;\gdelta{(+28)} \\
      \bottomrule
    \end{tabular}%
  }}
\end{table}

\subsubsection{Results on General Tasks}
As shown in Tab.~\ref{tab:real}(a), MemoryVLA achieves an average success score of 85\% on real-robot general tasks, outperforming CogACT by +9 points. It improves over the strong baseline on all 6 tasks, with notable gains on \textit{Egg in Pan} (+13) and \textit{Egg in Oven} (+20). 

\subsubsection{Results on Long-Horizon Memory-Dependent Tasks}
As shown in Tab.~\ref{tab:real}(b), MemoryVLA achieves an average success score of 83\% on long-horizon memory-dependent tasks, exceeding CogACT by +26 points. 
The improvements are larger than those on general tasks, including +43 on \textit{Seq. Push Buttons}, +38 on \textit{Change Food}, +32 on \textit{Guess Where}, and +17 on \textit{Clean Table \& Count}. 
These results indicate that explicit memory modeling is particularly beneficial for real-world tasks requiring long-term temporal context. 

\subsubsection{Results on Long-Horizon Imagination-Dependent Tasks}
As shown in Tab.~\ref{tab:real}(c), MemoryVLA++ achieves an average success score of 77\% on long-horizon imagination-dependent tasks, outperforming CogACT by +28 points. 
Compared with MemoryVLA, MemoryVLA++ further improves the average score by +12 points. 
The improvements are especially large on \textit{Bag Pack \& Zip} (+36) and \textit{Conveyor Scan-Pick} (+33). 
These results show that future imagination improves real-world tasks requiring future-state anticipation. 

\begin{figure*}[t]
    \centering
    \includegraphics[width=0.8\linewidth]{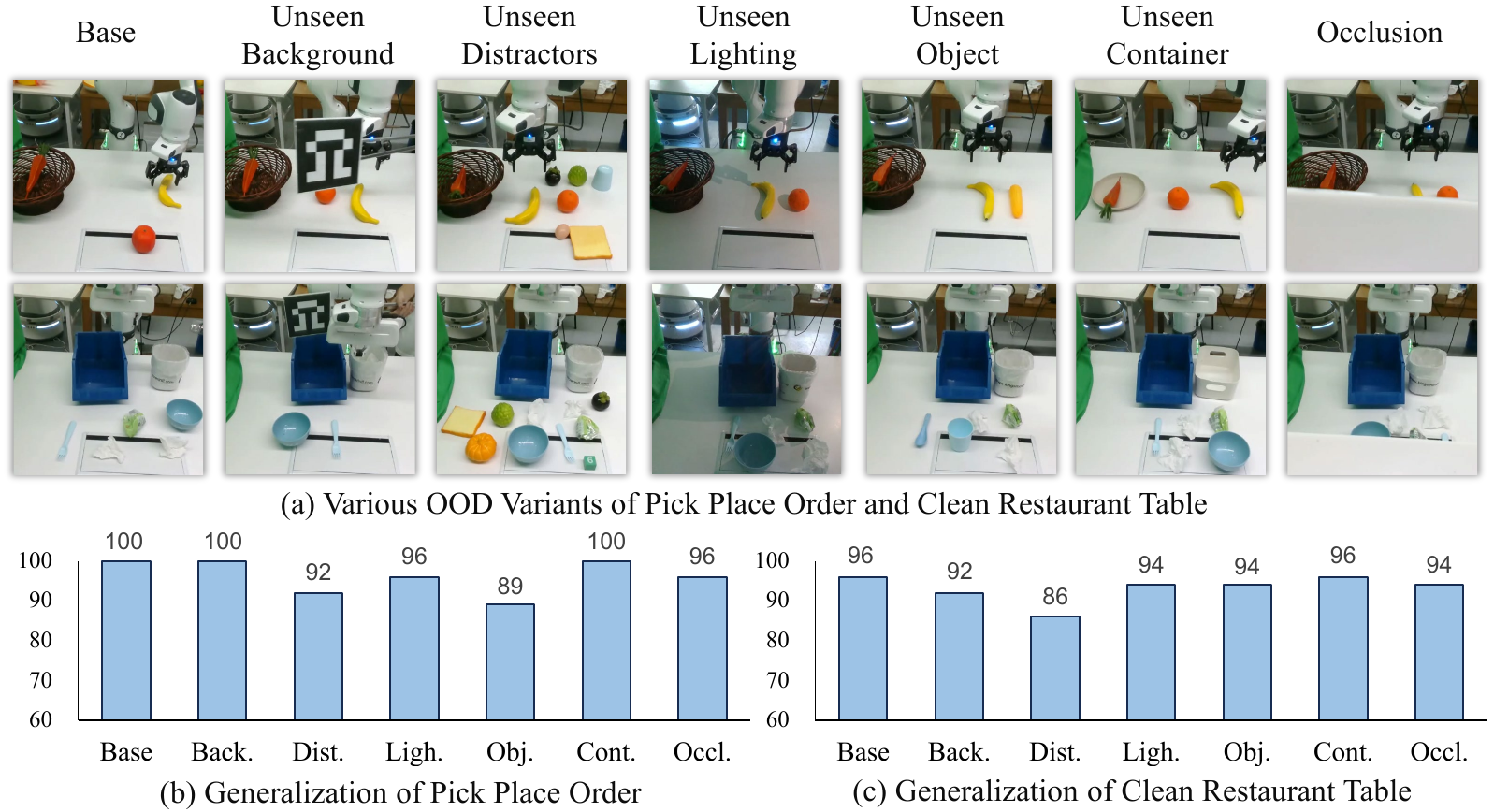}
    \caption{Real-world robustness and generalization under OOD conditions. Our method maintains strong performance across diverse variations.}
    \label{fig:ood_real}
\end{figure*}

\subsubsection{Results on Robustness and Generalization}
We further evaluate real-world robustness and generalization under diverse out-of-distribution conditions.
As shown in Fig.~\ref{fig:ood_real}, we test two long-horizon tasks, \textit{Pick Place Order} and \textit{Clean Restaurant Table}, under variations in backgrounds, distractors, lighting, objects, containers, and occlusions. 
The model shows only minor performance drops across these variations, indicating strong robustness to real-world distribution shifts. 

\subsection{Ablation Studies}
\label{sec:ablation}

\subsubsection{Ablation Studies On Memory Modeling}
\begin{table}[t]
  \centering
  \caption{Ablation studies on memory modeling. For memory length, Small/Default/Large denote 4/16/64 for SimplerEnv, 8/16/32 for Libero-Long-90, and 64/256/512 for Real-Temporal. Real-Temporal refers to the Clean Table \& Count task. The default setting is highlighted with \colorbox{gray!15}{Gray}.}
  \label{tab:ablation_mem}

  \subfloat[Memory length.]{
    \makebox[\linewidth][c]{%
    \begin{tabular}{cccc}
      \toprule
      Length & SimplerEnv & Long-90 & Real-Temporal \\
      \midrule
      Small   & 67.7 & 94.2 & 78 \\
      \rowcolor{gray!15}
      Default & \textbf{71.9} & \textbf{95.6} & \textbf{84} \\
      Large   & 67.7 & 95.6 & 81 \\
      \bottomrule
    \end{tabular}%
    }
  }\\[6pt]

  \subfloat[Memory fusion strategy.]{
    \makebox[\linewidth][c]{%
    \begin{tabular}{cccc}
      \toprule
      Fusion & SimplerEnv & Long-90 & Real-Temporal \\
      \midrule
      Add  & 67.7 & 93.8 & 78 \\
      \rowcolor{gray!15}
      Gate & \textbf{71.9} & \textbf{95.6} & \textbf{84} \\
      \bottomrule
    \end{tabular}%
    }
  }\\[6pt]

  \subfloat[Memory consolidation strategy.]{
    \makebox[\linewidth][c]{%
    \begin{tabular}{cccc}
      \toprule
      Consolidation & SimplerEnv & Long-90 & Real-Temporal \\
      \midrule
      FIFO        & 66.7 & 94.9 & 76 \\
      \rowcolor{gray!15}
      Token Merge & \textbf{71.9} & \textbf{95.6} & \textbf{84} \\
      \bottomrule
    \end{tabular}%
    }
  }\\[6pt]

  \makebox[\linewidth][c]{%
  \subfloat[Memory retrieval strategy.]{
    \resizebox{0.46\linewidth}{!}{%
    \begin{tabular}{cc}
      \toprule
      Retrieval & SimplerEnv \\
      \midrule
      w/o Timestep PE & 69.8 \\
      \rowcolor{gray!15}
      w/ Timestep PE & \textbf{71.9} \\
      \bottomrule
    \end{tabular}%
    }
  }
  \hspace{0.03\linewidth}
  \subfloat[Memory type.]{
    \resizebox{0.46\linewidth}{!}{%
    \begin{tabular}{cc}
      \toprule
      Memory Type & SimplerEnv \\
      \midrule
      Cognitive Memory  & 63.5 \\
      Perceptual Memory & 64.6 \\
      \rowcolor{gray!15}
      Both & \textbf{71.9} \\
      \bottomrule
    \end{tabular}%
    }
  }}
\end{table}

We conduct ablations based on MemoryVLA to analyze the key designs of memory modeling in Tab.~\ref{tab:ablation_mem}. 
As shown in Tab.~\ref{tab:ablation_mem}(a), the default memory length achieves the best overall performance across SimplerEnv, Libero-Long-90, and the real-world temporal task, showing the importance of balancing temporal coverage and memory redundancy. 
As shown in Tab.~\ref{tab:ablation_mem}(b), gated fusion consistently outperforms addition-based fusion, demonstrating the benefit of adaptive memory integration. 
For memory consolidation, Token Merge consistently improves over FIFO, confirming the benefit of redundancy-aware memory compression, as shown in Tab.~\ref{tab:ablation_mem}(c). 
As shown in Tab.~\ref{tab:ablation_mem}(d), timestep positional encoding further improves memory retrieval by preserving temporal order. 
As shown in Tab.~\ref{tab:ablation_mem}(e), combining perceptual and cognitive memory performs best, outperforming either memory type alone by a clear margin. 
Overall, these ablations confirm the effectiveness of the proposed temporal memory modeling design. 

\subsubsection{Ablation Studies On Imagination Modeling}
We conduct ablations based on MemoryVLA++ to analyze the key designs of imagination modeling in Tab.~\ref{tab:ablation_imag}. 
As shown in Tab.~\ref{tab:ablation_imag}(a), using a single denoising step already achieves competitive performance, while more steps bring little gain and increase inference cost. 
As shown in Tab.~\ref{tab:ablation_imag}(b), a longer imagination horizon improves performance, suggesting that future temporal cues are useful for long-horizon decision-making. 
For world-model update, freezing the world model outperforms updating it during policy training, as shown in Tab.~\ref{tab:ablation_imag}(c), indicating that the pretrained visual dynamics prior should be preserved. 
As shown in Tab.~\ref{tab:ablation_imag}(d), memory-guided integration clearly outperforms direct addition, showing the importance of selecting decision-relevant imagined cues through memory context. 
Overall, these ablations confirm the effectiveness of the proposed imagination modeling design.

\begin{table}[t]
  \centering
  \caption{Ablation studies on imagination modeling. Success rates (\%) are reported on Mikasa-Robo. The default setting is highlighted with \colorbox{gray!15}{Gray}.}
  \label{tab:ablation_imag}

  \makebox[\linewidth][c]{%
  \subfloat[Denoising step.\label{tab:ablation_imag_step}]{
    \begin{minipage}{0.38\linewidth}
      \centering
      \begin{tabular}{cc}
        \toprule
        Denoise Step & Avg. Succ. \\
        \midrule
        \rowcolor{gray!15}
        1 & 44.4 \\
        3 & \textbf{44.6} \\
        5 & 43.6 \\
        \bottomrule
      \end{tabular}
    \end{minipage}
  }
  \hspace{0.1\linewidth}
  \subfloat[Imagination horizon.\label{tab:ablation_imag_horizon}]{
    \begin{minipage}{0.38\linewidth}
      \centering
      \begin{tabular}{cc}
        \toprule
        Horizon & Avg. Succ. \\
        \midrule
        4  & 43.4 \\
        8  & 43.8 \\
        \rowcolor{gray!15}
        16 & \textbf{44.4} \\
        \bottomrule
      \end{tabular}
    \end{minipage}
  }%
  }\\[6pt]

  \makebox[\linewidth][c]{%
  \subfloat[World-model update.\label{tab:ablation_imag_freeze}]{
    \begin{minipage}{0.38\linewidth}
      \centering
      \begin{tabular}{cc}
        \toprule
        WM Update & Avg. Succ. \\
        \midrule
        w/o Freeze & 42.8 \\
        \rowcolor{gray!15}
        w/ Freeze & \textbf{44.4} \\
        \bottomrule
      \end{tabular}
    \end{minipage}
  }
  \hspace{0.06\linewidth}
  \subfloat[Integration strategy.\label{tab:ablation_integration}]{
    \begin{minipage}{0.38\linewidth}
      \centering
      \begin{tabular}{cc}
        \toprule
        Integration & Avg. Succ. \\
        \midrule
        Add & 41.2 \\
        \rowcolor{gray!15}
        Mem-Guided & \textbf{44.4} \\
        \bottomrule
      \end{tabular}
    \end{minipage}
  }%
  }
\end{table}

\subsection{Analytical Results}
\label{sec:analysis}

\begin{figure*}[t]
    \centering
    \includegraphics[width=0.8\linewidth]{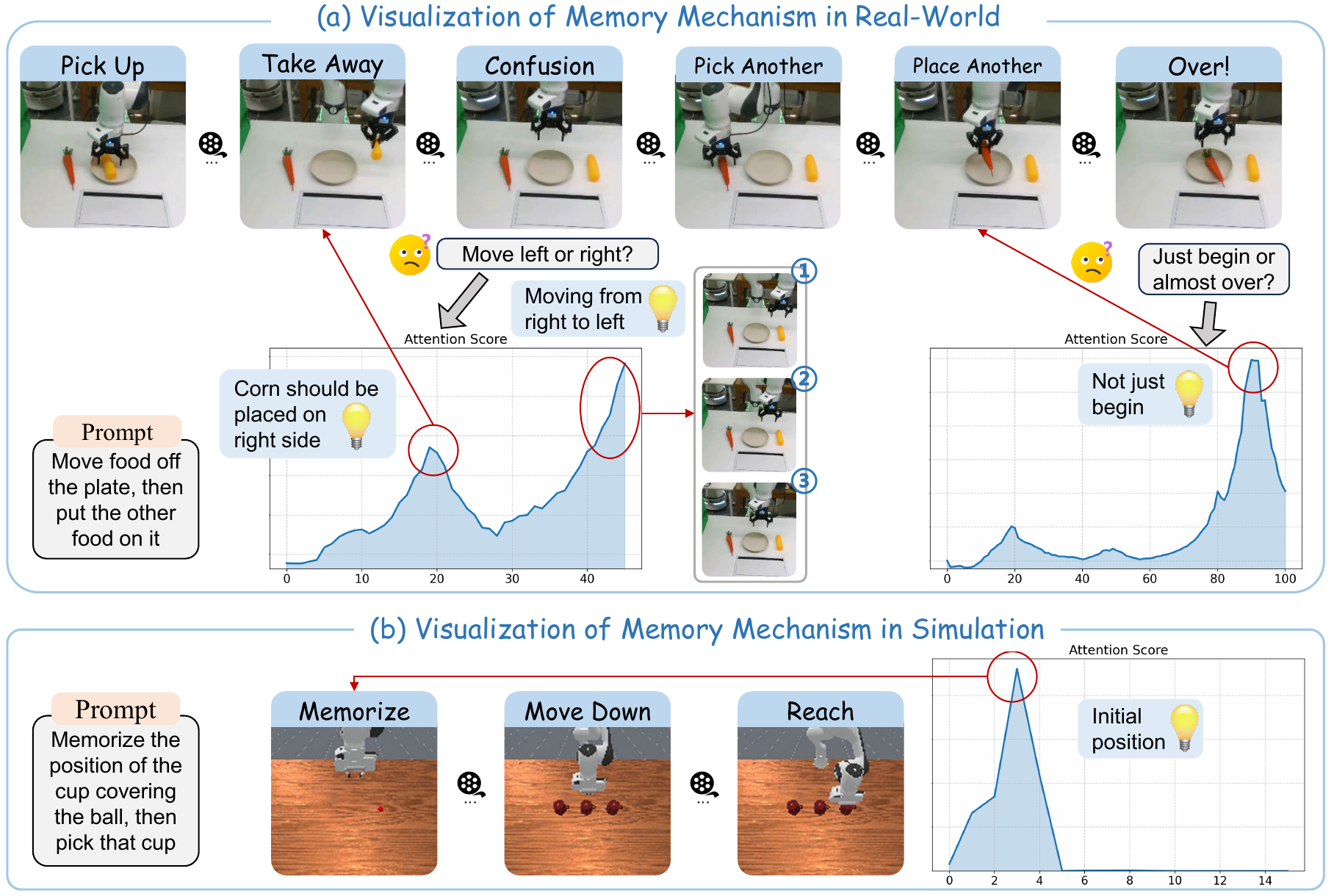}
    \caption{Analysis of memory modeling in real-world and simulation tasks.
    We visualize the retrieved memory elements and their attention weights in the real-world \textit{Change Food} task and the simulated \textit{Shell Game Touch} task, showing that the model attends to decision-relevant past frames that resolve ambiguities absent from the current observation.}
    \label{fig:mem_attn}
\end{figure*}

\subsubsection{Analysis of Memory Mechanism}
To provide a direct view of how the memory mechanism functions, Fig.~\ref{fig:mem_attn} visualizes the retrieved memory elements and their attention weights on the real-world and simulation tasks. The model consistently attends to past frames that resolve decision-relevant ambiguities absent from the current observation.
In the real-world \textit{Change Food} task, after the first food item is placed aside, the current frame contains two food items on the table, making it impossible to determine from this single observation which one should be picked next. Our method therefore attends strongly to the nearby frames reflecting the recent motion trend, as well as the last decisive frame before the ambiguity arises. 
In the \textit{Shell Game Touch} task from Mikasa-Robo Simulation Benchmark, the robot is briefly shown the cube location before it is covered by cups. The model consistently attends to the initial revealing frames, which provide the only reliable cue for identifying the correct cup. These results demonstrate that our method retrieves meaningful temporal cues essential for disambiguating the next action, rather than simply recalling redundant visual history. 

\subsubsection{Analysis of Imagination Mechanism}

\begin{figure*}[h!]
    \centering
    \includegraphics[width=1\linewidth]{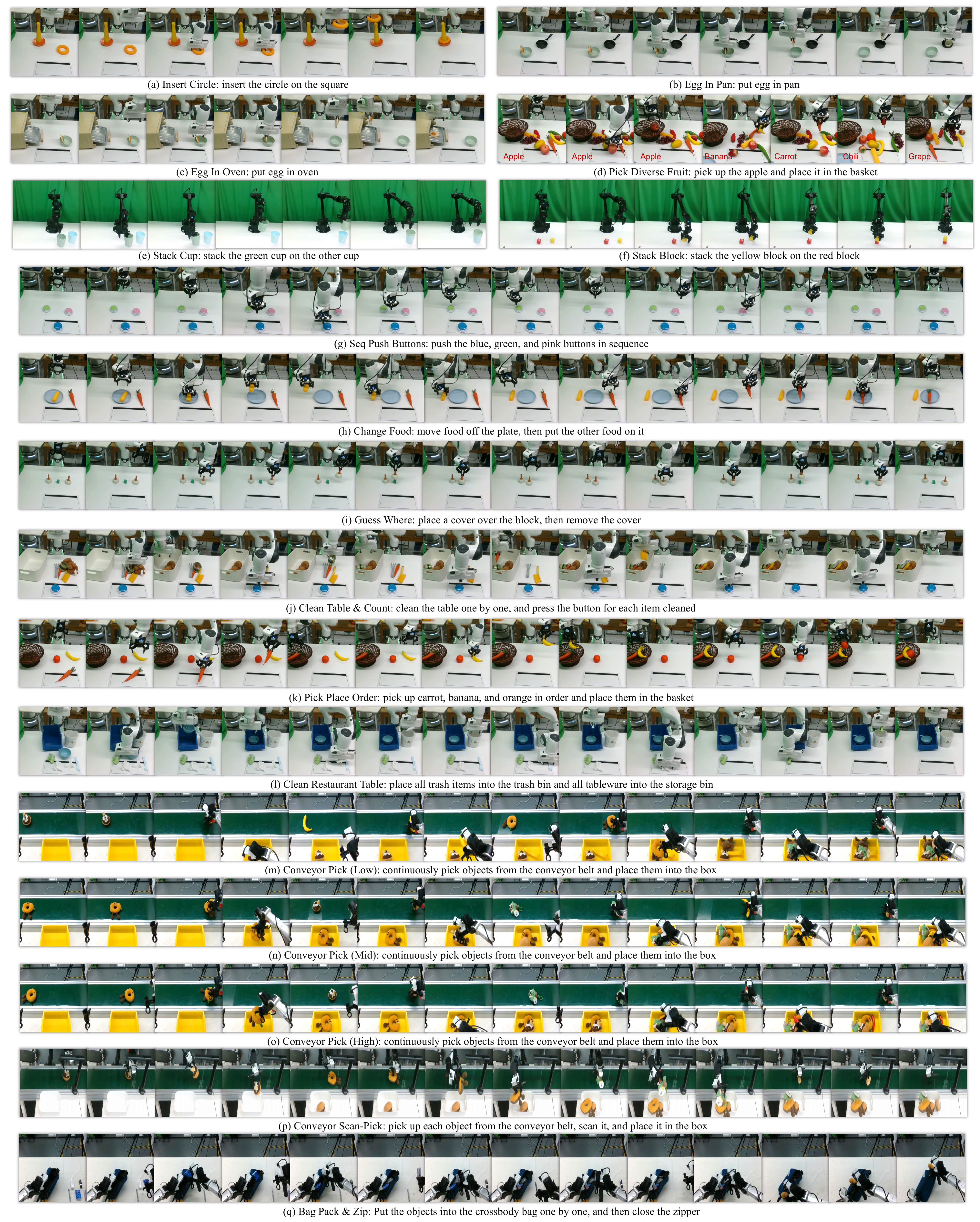}
    \caption{Visualization of real-robot tasks across general manipulation, long-horizon memory-dependent tasks, and long-horizon imagination-dependent tasks.}
    \label{fig:vis_real}
\end{figure*}

\begin{figure*}[t]
    \centering
    \includegraphics[width=1\linewidth]{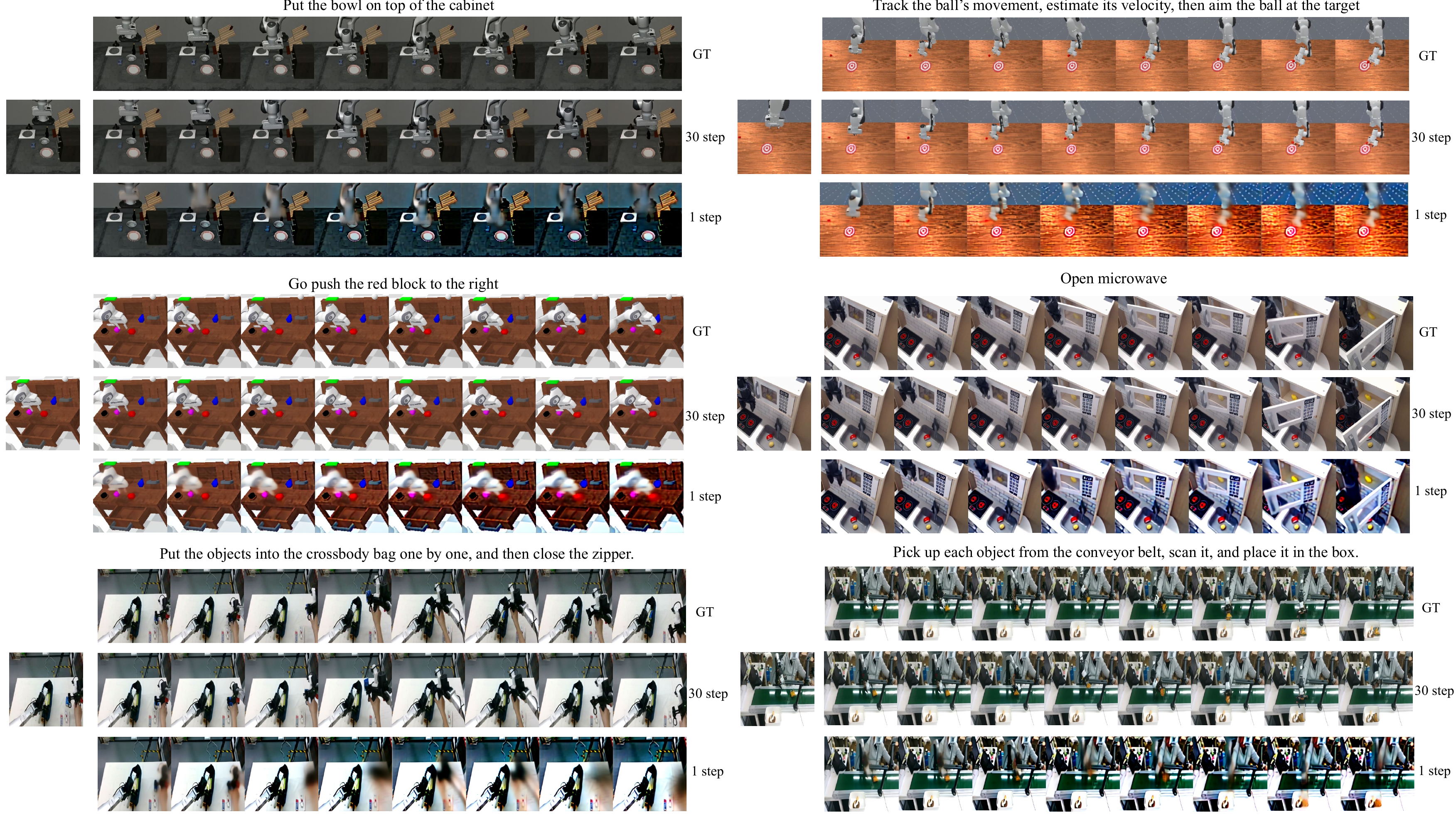}
    \caption{Visualization of world model imagination. 
    We compare ground-truth future observations with imagined trajectories generated using 30 and 1 denoising steps across simulation and real-world tasks, showing that the world model captures future semantic dynamics even under partial denoising.}
    \label{fig:vis_world}
\end{figure*}

\begin{table}[t]
\centering
\caption{Evaluation metrics for simulation and real-robot datasets. PSNR and SSIM are higher-is-better ($\uparrow$), while LPIPS, FVD, and EPE are lower-is-better ($\downarrow$).}
\label{tab:world_model}
\resizebox{1.0\linewidth}{!}{
\begin{tabular}{lccccc}
\toprule
Datasets & PSNR $\uparrow$ & SSIM $\uparrow$ & LPIPS $\downarrow$ & FVD $\downarrow$ & EPE $\downarrow$ \\
\midrule
Libero & 20.26 & 0.820 & 0.182 & 101.93 & 0.5104 \\
Bridge & 17.44 & 0.712 & 0.276 & 132.31 & 1.4999 \\
Mikasa-Robo & 26.39 & 0.838 & 0.174 & 189.38 & 0.1540 \\
Calvin & 22.22 & 0.833 & 0.185 & 29.69 & 0.2049 \\
Real-Conv. Pick & 16.91 & 0.764 & 0.250 & 128.18 & 1.6276 \\
Real-Conv. Scan & 22.34 & 0.822 & 0.182 & 82.92 & 0.4163 \\
Real-Bag Pack & 16.94 & 0.768 & 0.266 & 70.56 & 1.7672 \\
\midrule
Average & 20.36 & 0.794 & 0.216 & 105.00 & 0.8829 \\
\bottomrule
\end{tabular}
}
\end{table}

As shown in Fig.~\ref{fig:vis_world}, we visualize the imagined future trajectories from the world model and compare them with ground-truth future observations.
Across both simulation and real-world tasks, the generated trajectories preserve the main semantic evolution of the scene. Notably, using only 1 denoising step already captures meaningful future dynamics, supporting our latent-space imagination design for efficient robotic decision-making. 

Table~\ref{tab:world_model} quantitatively evaluates imagination quality across simulation and real-robot datasets under the full-denoising setting.
PSNR~\cite{hore2010image} measures pixel-level fidelity, SSIM~\cite{wang2004image} measures frame-level structural similarity, and LPIPS~\cite{zhang2018unreasonable} measures perceptual similarity with deep features. 
FVD~\cite{unterthiner2018towards} evaluates video-level distributional quality, while Flow-EPE~\cite{zhang2025reinforcing} measures motion consistency using optical-flow endpoint error. 
The results show that the world model captures future semantic and motion dynamics.

\subsubsection{Analysis of Inference Efficiency}
\begin{table*}[t]
  \centering
  \caption{Inference efficiency comparison. Latency, throughput, and GPU memory are measured over 300 runs in bfloat16 using single-view inputs and an action chunk length of 16 on RTX 4090 and HGX H20 GPUs.}
  \label{tab:efficiency}
  \resizebox{0.9\linewidth}{!}{%
  \begin{tabular}{lccccc}
    \toprule
    Method
      & Latency (RTX 4090)
      & Throughput (RTX 4090)
      & Latency (H20)
      & Throughput (H20)
      & GPU Memory \\
    \midrule
    Baseline & 0.187 s & 85.6 Hz & 0.236 s & 67.8 Hz & 15.8 GB \\
    MemoryVLA & 0.194 s & 82.5 Hz & 0.246 s & 65.0 Hz & 16.6 GB \\
    MemoryVLA++ & 0.241 s & 66.4 Hz & 0.301 s & 53.2 Hz & 21.7 GB \\
    \bottomrule
  \end{tabular}%
  }
\end{table*}

As shown in Tab.~\ref{tab:efficiency}, we compare inference latency, throughput, and GPU memory with the baseline on RTX 4090 and HGX H20 GPUs. 
All measurements are averaged over 300 runs in bfloat16 using single-view inputs and an action chunk length of 16. 
MemoryVLA introduces only minor overhead. On RTX 4090, the latency only increases from 0.187\,s to 0.194\,s, corresponding to only a \(\sim\)4\% increase, while the throughput remains close to the baseline, decreasing from 85.6\,Hz to 82.5\,Hz. GPU memory increases by only 0.8\,GB. These results show that the memory module is lightweight and introduces negligible extra cost during inference. 
MemoryVLA++ further introduces latent-space imagination during inference. 
Its latency increases to 0.241\,s on RTX 4090 and 0.301\,s on H20, while still maintaining 66.4\,Hz and 53.2\,Hz throughput, respectively. 
This moderate overhead comes from the additional world-model forward pass, but remains practical for real-robot deployment. 

\subsubsection{Analysis of Stronger VLA Pretraining}

As shown in Tab.~\ref{tab:backbone}, replacing the LLaMA2-based VLA with a Qwen2.5-based VLA and Dexbotic pretraining improves the average success rate from 71.9\% to 84.4\% on SimplerEnv. 
On Libero, the two settings achieve comparable performance, with Qwen2.5 and Dexbotic pretraining slightly improving the average success rate from 96.7\% to 97.0\%. 
These results indicate that stronger VLA pretraining can further improve performance, especially on more challenging tasks. 

\begin{table}[h]
  \centering
  \caption{Analysis of VLA backbone and pretraining.}
  \vspace{-1em}
  \label{tab:backbone}
  \scriptsize
  \setlength{\tabcolsep}{2.5pt}
  \renewcommand{\arraystretch}{1.05}

  \newcolumntype{Y}{>{\centering\arraybackslash}X}

  \subfloat[Performance on SimplerEnv.]{
    \makebox[\linewidth][c]{%
    \begin{tabularx}{0.95\linewidth}{>{\raggedright\arraybackslash}p{0.13\linewidth}
                                      >{\raggedright\arraybackslash}p{0.15\linewidth}
                                      YYYYY}
      \toprule
      Backbone & Pretraining
        & \makecell[c]{Spoon\\Towel}
        & \makecell[c]{Carrot\\Plate}
        & \makecell[c]{Stack\\Cube}
        & \makecell[c]{Eggplant\\Basket}
        & \makecell[c]{Avg.\\Succ.} \\
      \midrule
      LLaMA2  & CogACT   & 75.0  & 75.0 & 37.5 & 100.0 & 71.9 \\
      \rowcolor{gray!15}
      Qwen2.5 & Dexbotic & 100.0 & 66.7 & 70.8 & 100.0 & 84.4 \\
      \bottomrule
    \end{tabularx}%
    }
  }\\[6pt]

  \subfloat[Performance on Libero.]{
    \makebox[\linewidth][c]{%
    \begin{tabularx}{0.95\linewidth}{
      >{\raggedright\arraybackslash}p{0.13\linewidth}
      >{\raggedright\arraybackslash}p{0.15\linewidth}
      YYYY
      >{\centering\arraybackslash}p{0.13\linewidth}
    }
      \toprule
      Backbone & Pretraining
        & Spatial
        & Object
        & Goal
        & Long-10
        & \multicolumn{1}{c}{Avg. Succ.} \\
      \midrule
      LLaMA2  & CogACT   & 98.4 & 98.4 & 96.4 & 93.4 & 96.7 \\
      \rowcolor{gray!15}
      Qwen2.5 & Dexbotic & 97.2 & 99.2 & 98.4 & 93.2 & 97.0 \\
      \bottomrule
    \end{tabularx}%
    }
  }
\end{table}

\subsection{Visualization Analysis}
\label{sec:visual}

 Fig.~\ref{fig:vis_real} shows real-robot tasks across general manipulation, long-horizon memory-dependent tasks, and long-horizon imagination-dependent tasks.

\section{Conclusion}
\label{sec:conclusion}

We propose MemoryVLA++, a full temporal modeling framework that equips VLA models with memory and imagination for robotic manipulation. 
MemoryVLA++ leverages the commonsense priors of pretrained VLMs for high-level cognition. Inspired by cognitive science, it uses a hippocampus-like Perceptual-Cognitive Memory Bank that cooperates with working memory to capture past temporal dependencies. Meanwhile, a world model generates latent-space future imagination, and a full temporal-aware action expert uses memory and imagination to produce temporally consistent actions. 
Extensive experiments on 5 simulation benchmarks and 3 categories of real-robot tasks across 3 robots, covering nearly 200 tasks with diverse variations, demonstrate the effectiveness of MemoryVLA++. The improvements are especially clear on long-horizon temporal tasks, validating the importance of modeling past memory and future imagination for robotic manipulation. 
MemoryVLA++ represents an encouraging step toward temporal modeling in generalist robot policies.

\bibliographystyle{IEEEtran}
\bibliography{main}

\vspace{-22pt}
\begin{IEEEbiography}[{\includegraphics[width=1in,height=1.25in,clip]{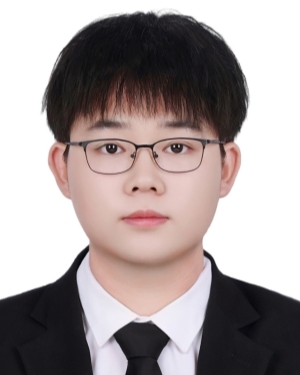}}]{Hao Shi}
received the BS degree in computer science from Tianjin University, Tianjin, China, in 2023. He is currently working toward the MS degree with the Department of Automation, Tsinghua University, Beijing, China. He is also a Visiting Student with MMLab, The University of Hong Kong, Hong Kong, China. His research interests include embodied intelligence and computer vision.
\end{IEEEbiography}
\vspace{-14pt}

\begin{IEEEbiography}[{\includegraphics[width=1in,height=1.25in,clip]{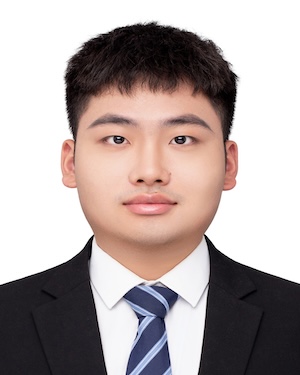}}]{Weiye Li}
received the BS degree in automation from Beihang University, Beijing, China, in 2025. He is currently working toward the MS degree with the Department of Automation, Tsinghua University, Beijing, China. His research interests include embodied intelligence.
\end{IEEEbiography}
\vspace{-14pt}

\begin{IEEEbiography}[{\includegraphics[width=1in,height=1.25in,clip]{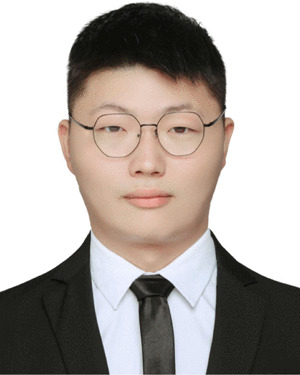}}]{Bin Xie}
received the MS degree in electronic information engineering from Tianjin University, Tianjin, China, in 2025. He is currently a Researcher with Dexmal. His research interests include embodied intelligence and computer vision.
\end{IEEEbiography}
\vspace{-14pt}

\begin{IEEEbiography}[{\includegraphics[width=1in,height=1.25in,clip]{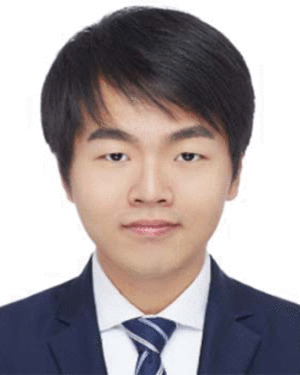}}]{Yulin Wang}
received the BS degree in automation from Beihang University, Beijing, China, in 2019, and the PhD degree in automation from Tsinghua University, Beijing, China, in 2025. He was a Visiting Student with the University of California, Berkeley, Berkeley, CA, USA, in 2018. He is currently a Postdoctoral Researcher with the University of Oxford, Oxford, U.K. His research interests include deep learning and computer vision.
\end{IEEEbiography}
\vspace{-14pt}

\begin{IEEEbiography}[{\includegraphics[width=1in,height=1.25in,clip]{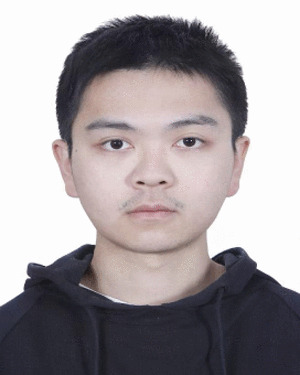}}]{Renping Zhou}
received the BS degree in computer science and technology in 2025 from Tsinghua University, Beijing, China. He is currently working toward the PhD degree with the Department of Automation, Tsinghua University, Beijing, China. His research interests include embodied intelligence and computer vision.
\end{IEEEbiography}
\vspace{-14pt}

\begin{IEEEbiography}[{\includegraphics[width=1in,height=1.25in,clip]{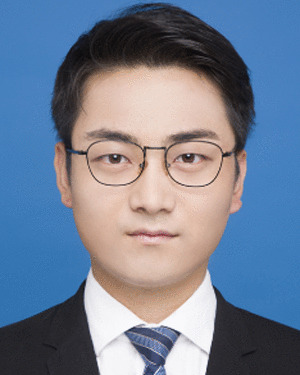}}]{Tiancai Wang}
received the MS degree from Tianjin University, Tianjin, China. He was a Research Intern with the Inception Institute of Artificial Intelligence, Abu Dhabi, United Arab Emirates, and a Senior Researcher with Megvii Research, Beijing, China. He is currently a Co-founder and Researcher with Dexmal. His research interests include embodied intelligence and computer vision.
\end{IEEEbiography}
\vspace{-14pt}

\begin{IEEEbiography}[{\includegraphics[width=1in,height=1.25in,clip]{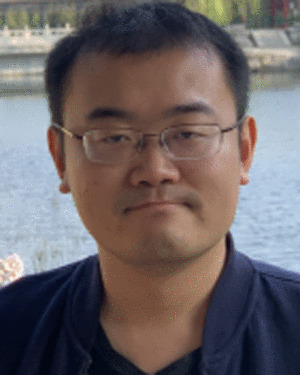}}]{Xiangyu Zhang}
received the PhD degree from Xi'an Jiaotong University, Xi'an, China, in 2017. He was a Principal Researcher with Megvii Research, Beijing, China. He is currently a Co-founder and Chief Scientist with Stepfun. His research interests include computer vision and deep learning. He got the CVPR Best Paper Award in 2016. The total number of his Google Scholar citations is more than 450,000.
\end{IEEEbiography}
\vspace{-14pt}

\begin{IEEEbiography}[{\includegraphics[width=1in,height=1.25in,clip]{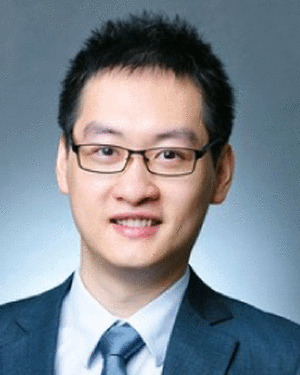}}]{Ping Luo}
(Senior Member, IEEE) received the PhD degree in information engineering from the Chinese University of Hong Kong, Hong Kong, in 2014. He is currently an Associate Professor with the Department of Computer Science, The University of Hong Kong, Hong Kong, China. From 2014 to 2016, he was a Postdoctoral Researcher with the Chinese University of Hong Kong. From 2017 to 2018, he was a Principal Research Scientist with SenseTime Research. His research interests include machine learning and computer vision.
\end{IEEEbiography}
\vspace{-14pt}

\begin{IEEEbiography}[{\includegraphics[width=1in,height=1.25in,clip]{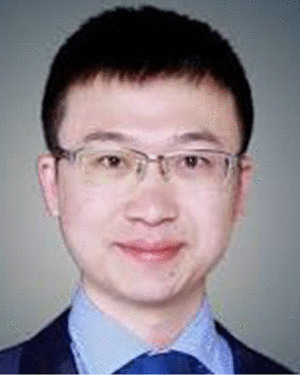}}]{Gao Huang}
(Member, IEEE) received the BS degree in automation from Beihang University, Beijing, China, in 2009, and the PhD degree in automation from Tsinghua University, Beijing, China, in 2015. He was a Visiting Research Scholar with the Department of Computer Science and Engineering, Washington University in St. Louis, St. Louis, MO, USA, in 2013, and a Postdoctoral Researcher with the Department of Computer Science, Cornell University, Ithaca, NY, USA, from 2015 to 2018. He is currently an Associate Professor with the Department of Automation, Tsinghua University, Beijing, China. His research interests include machine learning and deep learning.
\end{IEEEbiography}
\vspace{-14pt}

\vfill

\clearpage
\appendices

\phantomsection
\addcontentsline{toc}{section}{Supplementary Materials}
\section*{Supplementary Materials}

\subsection{Additional Simulation Training and Evaluation Setup}

\subsubsection{Libero}
Following OpenVLA~\cite{kim2025openvla}, we train with 50 demonstrations per task after removing failed trajectories from the dataset. 
For MemoryVLA++, Spatial, Object, Goal, and Long-10 are jointly trained for 60k steps to align with the subsequent Libero-Plus evaluation, since Libero-Plus is built on these four standard suites. 
Long-90 is trained separately for 30k steps because it is not included in Libero-Plus. 
The dataloader adopts a grouped sampling strategy in which each batch is divided into multiple groups, and each group consists of several frames drawn from a single episode. Frames within a group are kept in temporal order. The memory length is set to 16. 
The corresponding world model is trained for 40k steps. 
For MemoryVLA, we follow its original training protocol~\cite{shi2025memoryvla}, and we observe similar performance between separate and joint training on the Libero.

Evaluation on Libero~\cite{liu2023libero} is conducted across all five suites: Spatial, Object, Goal, Long-10, and Long-90. 
For Spatial, Object, Goal, and Long-10, evaluation is performed every 10k steps. 
For Long-90, evaluation is performed every 2k steps. 
Each task is evaluated with 50 trials, and success rates are reported at the best evaluation checkpoint. 
Unlike SimplerEnv, no action ensemble strategy is used in our Libero experiments. 
CogACT results are reproduced using the official codebase for fairness. 

\subsubsection{SimplerEnv}
On BridgeData-v2, models are trained for 50k steps with a stream dataloader. 
Each episode is unpacked into consecutive frames tagged with its episode ID. 
During training, batches are filled sequentially with frames from a single episode whenever possible. 
If an episode ends before the batch is complete, the remaining slots are filled with frames from the following episode. 
A new batch then continues from the position where the previous one stopped, ensuring that in-episode temporal order is always preserved. 
The memory length is fixed to 16. 
The corresponding world model is trained for 40k steps.

Evaluation follows the official CogACT protocol~\cite{li2024cogact}. 
We adopt the same evaluation scripts and use the adaptive action ensemble strategy introduced in CogACT, with ensemble coefficient $\alpha=0.1$ and ensemble horizon set to $7$ for Bridge. 
Models are validated every 2.5k steps. 
Since the denoising objective of diffusion models does not reliably indicate policy quality, we report success rates at the best validation checkpoint. 
Each task is evaluated over 24 trials. 

Since the original paper only reported per-task success rates for CogACT-Base but not for CogACT-Large, we re-evaluated the released CogACT-Large checkpoint in our setup and report those numbers for fairness. 
For $\pi_0$, results are taken from the open-source reproduction \href{https://github.com/allenzren/open-pi-zero}{open-pi-zero}, which provides implementations with both \textit{uniform} and \textit{beta} timestep sampling strategies in flow matching. We report results under float32 precision as in the public release.

\subsubsection{Mikasa-Robo}
For Mikasa-Robo~\cite{cherepanov2025memory}, we adopt the standard protocol with five tasks and train jointly on all 1,250 demonstrations for 20k steps, using $128\times128$ RGB observations and $\Delta$ end-effector control. 
We reuse the same dataloader setup as in Libero, and set the memory length to 16. 
The corresponding world model is trained for 20k steps.

Validation is performed every 1k training steps, and success rates are reported at the checkpoint with the best validation performance. 
Each task is evaluated with 100 episodes. 
As in Libero, we do not use any action ensemble strategy in our Mikasa-Robo experiments.

\subsubsection{Calvin}
For Calvin~\cite{mees2022calvin}, we adopt the ABC$\rightarrow$D zero-shot setting, where policies are trained on environments A, B, and C and evaluated on the unseen environment D. We use 17,870 training trajectories from environments A, B, and C. The dataloader follows the same grouped sampling strategy as Libero.
Models are trained for 60k steps.
For the world model, we directly follow VPP~\cite{hu2024video}.

Models are evaluated every 2k steps and reported at the best evaluation checkpoint.
We report success rates over 1000 rollouts and the average number of completed tasks.

\subsubsection{Libero-Plus}
For Libero-Plus~\cite{fei2025libero}, we evaluate both zero-shot and supervised fine-tuning settings.
Libero-Plus introduces 7 out-of-distribution variations, covering camera, robot, language, lighting, background, noise, and layout.
In the zero-shot setting, models are trained on the 4 standard Libero suites, including Spatial, Object, Goal, and Long-10, and evaluated on Libero-Plus without using its training data.
In the supervised fine-tuning setting, Libero-Plus data are mixed with the 4 standard Libero suites for joint training.
Both settings use the same grouped dataloader as Libero and are trained for 60k steps.
OpenVLA-OFT and our method are both jointly trained on all suites and use wrist-view images in this benchmark.
The corresponding world model is trained for 40k steps.

Evaluation is performed every 10k steps, and results are reported at the best evaluation checkpoint.

\subsection{Additional Real-Robot Training and Evaluation Setup}

Models are trained for 5k-30k steps depending on task complexity and dataset size. 
The general tasks contain 50-150 demonstrations per task, while long-horizon memory-dependent and imagination-dependent tasks use 200-300 demonstrations per task. 
The memory length is set to 16 for general tasks and 256 for long-horizon temporal tasks. 
For MemoryVLA++, the world model is trained for 20k steps and used for long-horizon imagination-dependent tasks.

Evaluation uses 15-25 trials for General tasks and 10-15 trials for Long-horizon Temporal tasks.  
For General tasks, \textit{Pick Diverse Fruits} contains five variants (apple, orange, banana, chili, grape), each evaluated with 5 trials (25 total). 
All other General tasks are evaluated with 15 trials each, and we report task-level success rates.  

For long-horizon memory-dependent tasks, \textit{Seq. Push Buttons} includes three button orders (blue-pink-green, blue-green-pink, green-blue-pink), each tested with 5 trials. All other tasks are evaluated with 10 trials, and step-wise scoring is adopted to capture partial progress. The scoring rules are as follows: 
\begin{itemize}[leftmargin=*, nosep]
    \item \textbf{Seq. Push Buttons:} pressing each correct button yields 30, with a bonus of 10 if all three are correct. Loose matching is allowed (slight contact counts as a press). 
    \item \textbf{Change Food:} lifting and removing the initial food (30), grasping the new food (30), and placing it on the plate (30), with a 10 bonus for full success. 
    \item \textbf{Guess Where:} grasping the cover (30), covering the block (30), and uncovering it (40). 
    \item \textbf{Clean Table \& Count:} five objects in total. For each object, clearing yields 10 points and pressing the counter yields 10. Small counting errors (incomplete press / one extra press) earn 5; major errors (missed count / multiple extras) earn 0. Empty grasps with clear counting intent incur a 5-point penalty. 
    \item \textbf{Pick Place Order:} carrot, banana, and orange must be picked and placed in sequence. Each correct step earns 30, with a 10 bonus for full completion. Any order violation terminates the attempt. 
    \item \textbf{Clean Restaurant Table:} five objects in total. Each correctly sorted into trash bin or storage bin scores 20. Misplacement earns 10, and merely lifting without correct placement earns 5.  
\end{itemize}

For long-horizon imagination-dependent tasks, evaluation uses 10 consecutive trials per task. Step-wise scoring is adopted to capture partial progress within each trial. The scoring rules are as follows:
\begin{itemize}[leftmargin=*, nosep]
    \item \textbf{Conveyor-Pick:} five objects in total. Low-, mid-, and high-speed settings use the same scoring rule. For each object, successful grasping yields 10 points and placing it into the designated box yields 10.
    \item \textbf{Conveyor Scan-Pick:} five objects in total. For each object, successful grasping yields 6 points, successful scanning yields 6, and placing it into the box yields 6. A 2-point bonus is awarded if the object is fully completed.
    \item \textbf{Bag Pack \& Zip:} reaching into the bag with the left arm yields 10 points, opening the bag yields 10, and placing each of the five objects into the bag yields 10. A 10-point bonus is awarded if all five objects are successfully packed. Grasping the zipper pull with the right arm yields 10, and fully closing the zipper yields 10.
\end{itemize}

\subsection{Data Augmentation}
We apply standard per-frame augmentations to the third-person RGB stream during training. 
Augmentations are applied in a fixed order: random resized crop, random brightness, random contrast, random saturation, and random hue. 
The crop samples $90\%$ of the image area with aspect ratio $1.0$ and resizes to $224\times224$. 
Brightness is perturbed with magnitude $0.2$, contrast and saturation are scaled in $[0.8,\,1.2]$, and hue is shifted by up to $0.05$. 
All augmentations are disabled at evaluation.

\subsection{Data Length Statistics}

Tab.~\ref{tab:action_len} reports the maximum, minimum, median, and average action lengths across all task suites, including SimplerEnv Evaluation (Bridge, Fractal), LIBERO (Spatial/Object/Goal and 10/90 task suites), and both real-world general and temporal tasks. For the real-world tasks, we additionally provide filtered statistics based on a motion-magnitude threshold (translation $>$ 1 cm or rotation $>$ 0.4 rad between consecutive frames) to remove frames where the end-effector motion is small.

\begin{table}[h]
  \centering
  \caption{%
    Action length statistics for simulation and real-world task suites.
    For Calvin, we report the length of a 5-subtask sequence by multiplying the per-subtask action length by 5.
    For real-world tasks, ``Filtered'' removes frames with negligible end-effector motion
    (translation $<1$\,cm and rotation $<0.4$\,rad).
  }
  \label{tab:action_len}

  \begin{tabular}{lcccc}
    \toprule
    Task Suite & Max & Min & Median & Average \\
    \midrule
    Libero-Spatial / Object / Goal & 270  & 75  & 131 & 130 \\
    Libero-10 / 90                 & 505  & 58  & 144 & 156 \\
    SimplerEnv    & 200  & 80  & 117 & 119 \\
    Mikasa-Robo & 90 & 60 & 60 & 72 \\
    Calvin & 325 & 170 & 325 & 300 \\
    Libero-Plus & 505 & 75 & 138 & 156 \\
    Real-General (Original)        & 1575 & 281 & 575 & 575 \\
    Real-General (Filtered)        & 213  & 40  & 81  & 84  \\
    Real-Memory (Original)       & 7704 & 412 & 981 & 1672 \\
    Real-Memory (Filtered)       & 902  & 72  & 236 & 288 \\
    Real-Imagination (Original)   & 5737 & 1796 & 2809 & 2935 \\
    Real-Imagination (Filtered)   & 987 & 258 & 467 & 463 \\
    \bottomrule
  \end{tabular}
\end{table}

\subsection{Real-world Tasks Details}
Our real-world evaluation includes 17 tasks, divided into \textit{General}, \textit{Long-horizon Memory-dependent}, and \textit{Long-horizon Imagination-dependent} suites.

\paragraph{General Tasks.}  
\textit{Insert Circle}: insert a circle onto a vertical pillar, requiring accurate positioning and insertion. 

\textit{Egg in Pan}: place an egg into a shallow frying pan, testing grasp stability and gentle placement.  

\textit{Egg in Oven}: put an egg into a small oven container, involving more constrained placement than the pan. 

\textit{Stack Cups}: stack one plastic cup on top of another, evaluating vertical alignment and balance.  

\textit{Stack Blocks}: stack a yellow block on top of a red block, focusing on precise spatial alignment. 

\textit{Pick Diverse Fruits}: pick a specified fruit from a tabletop with more than ten different fruit types and place it into a basket, testing semantic understanding, visual diversity, and instruction following. 

\paragraph{Long-horizon Temporal Tasks.}  
\textit{Seq. Push Buttons}: push three buttons in a specified color sequence, stressing ordered memory and resistance to temporal confusion.  

\textit{Change Food}: remove a food item from a plate and replace it with another, requiring multi-step sequencing and correct temporal ordering.  

\textit{Guess Where}: cover a block with a container and later uncover it, testing reversible actions and consistent tracking over time.  

\textit{Clean Table \& Count}: clear items from the table one by one while pressing a counter button after each removal, combining manipulation with explicit progress monitoring.  

\textit{Pick Place Order}: pick up carrot, banana, and orange in a fixed order and place them into a basket, enforcing sequence-sensitive planning under temporal dependencies.  

\textit{Clean Restaurant Table}: sort table items by category, placing trash into a trash bin and tableware into a storage bin, representing a long-horizon task with semantic reasoning and complex multi-stage sequencing.  

\textit{Conveyor-Pick}: pick objects arriving sequentially on a moving conveyor belt and place them into a designated box, requiring anticipatory prediction of future object positions and stable grasping under continuous motion.  

\textit{Conveyor Scan-Pick}: pick objects from the conveyor belt, place each object under a scanner for identification, and then deposit it into a box, requiring anticipatory prediction for moving objects and memory for scan across the subsequent placement stage. 

\textit{Bag Pack \& Zip}: grasp objects one by one from the table, place them into a deformable bag, and finally grasp the zipper pull to close the bag, evaluating long-horizon planning, bimanual coordination, deformable-object manipulation, and fine-grained zipper alignment.

\subsection{Additional Qualitative Results}
\begin{figure*}[t]
\centering
\includegraphics[width=0.79\linewidth]{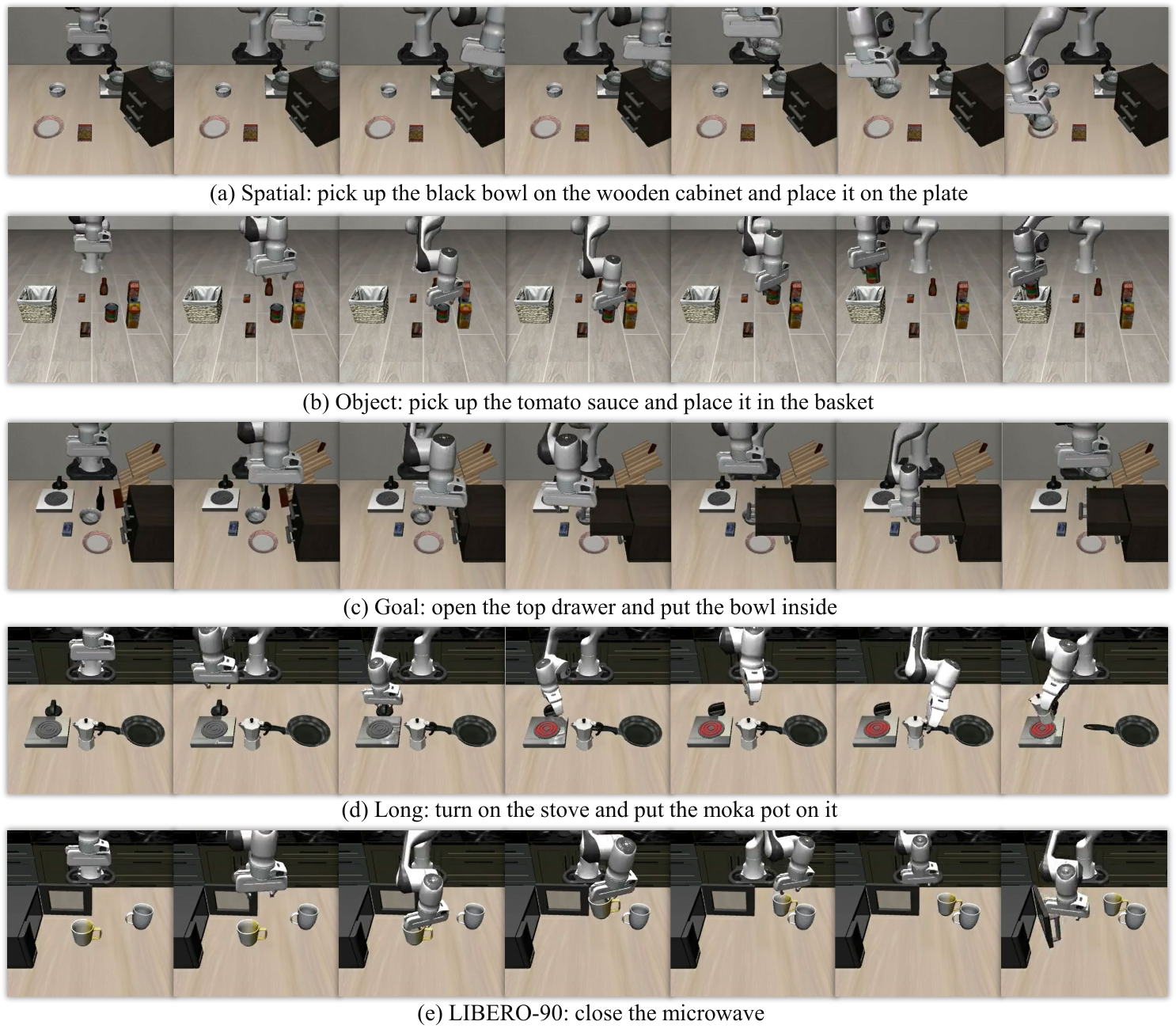}
\caption{Qualitative results on Libero.}
\label{fig:supp_libero}
\end{figure*}

\begin{figure*}[t]
\centering
\includegraphics[width=0.77\linewidth]{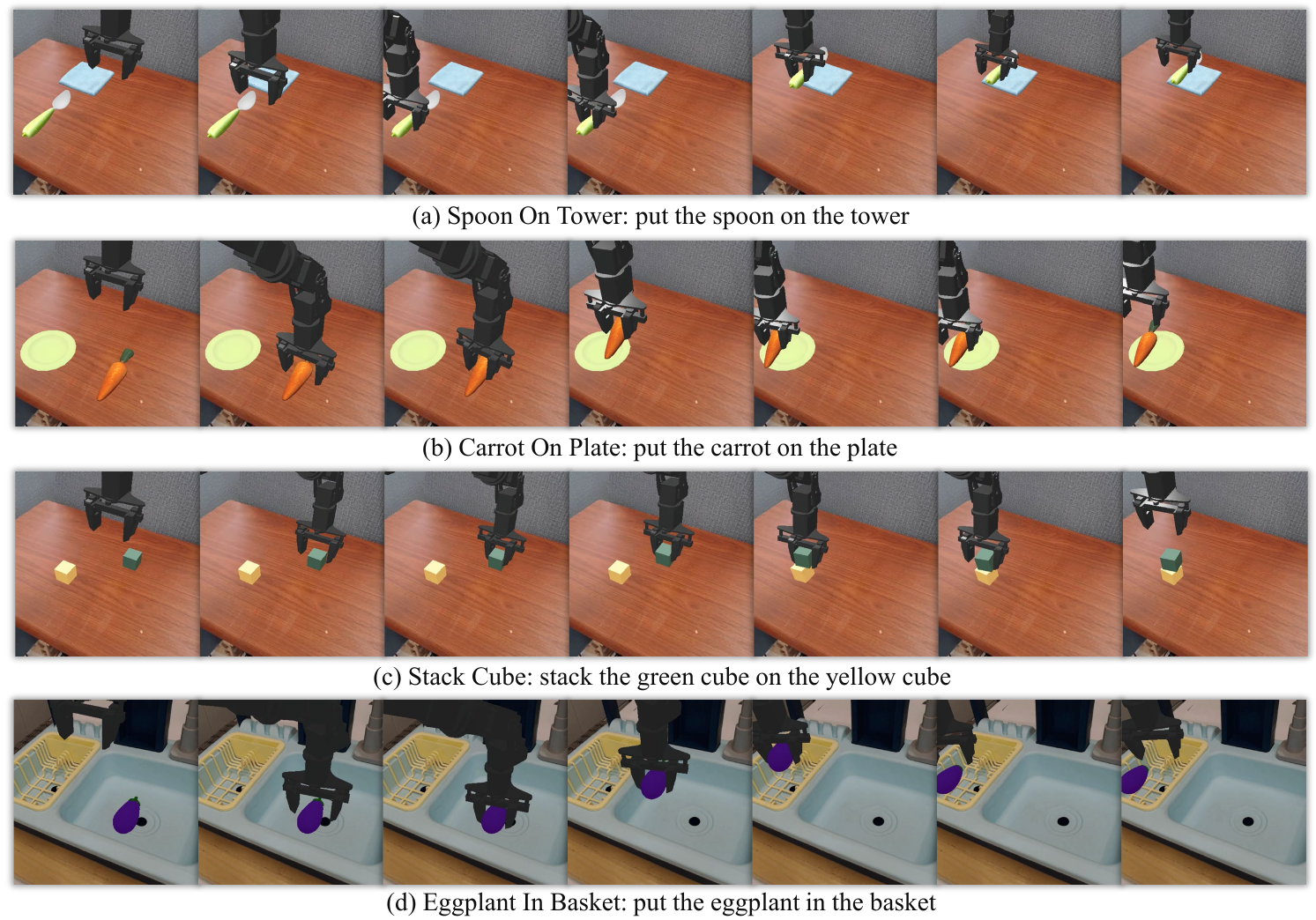}
\caption{Qualitative results on SimplerEnv.}
\label{fig:supp_simpler}
\end{figure*}

\begin{figure*}[t]
\centering
\includegraphics[width=0.8\linewidth]{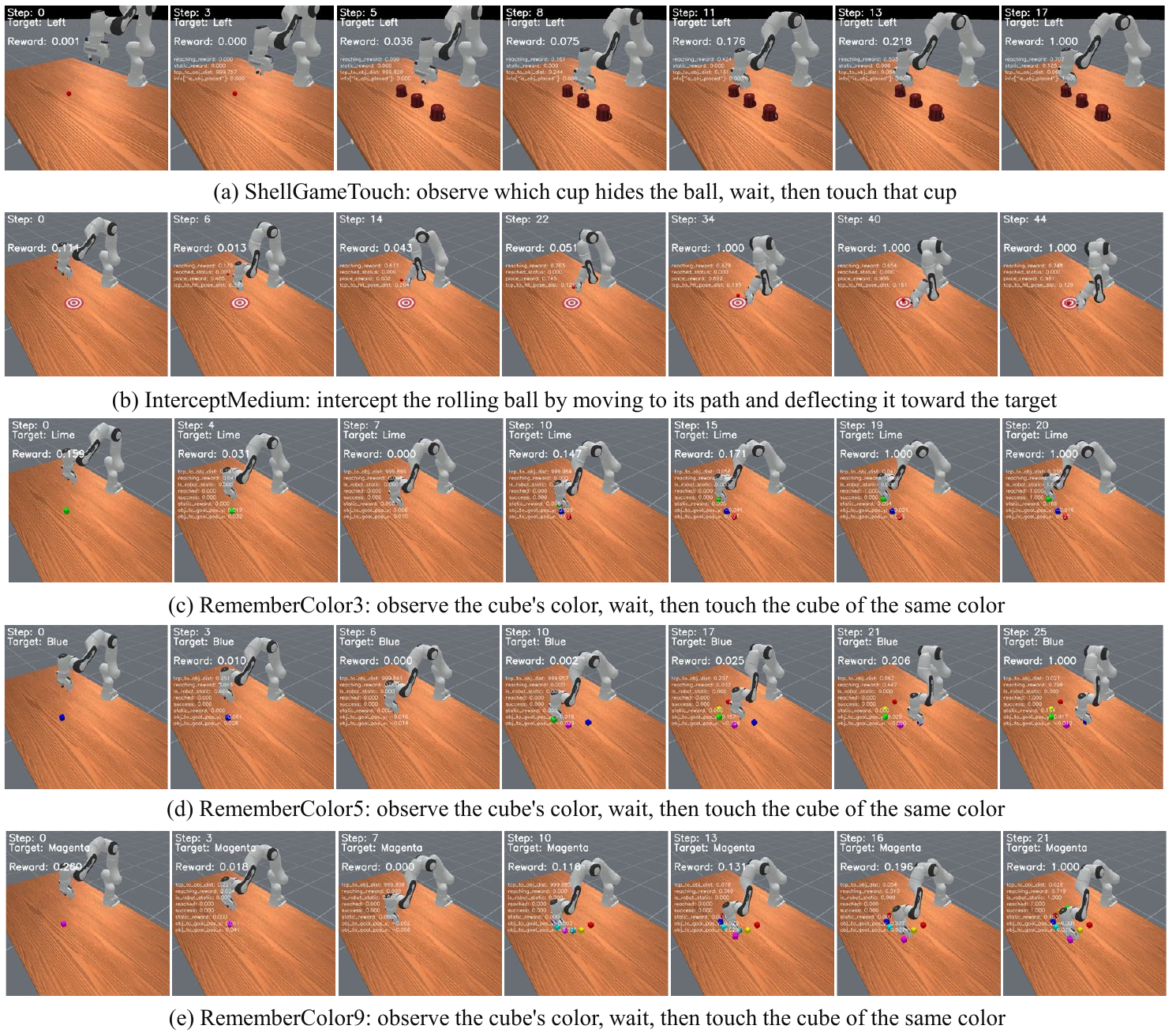}
\caption{Qualitative results on Mikasa-Robo.}
\label{fig:supp_mikasa}
\end{figure*}

\begin{figure*}[t]
\centering
\includegraphics[width=0.8\linewidth]{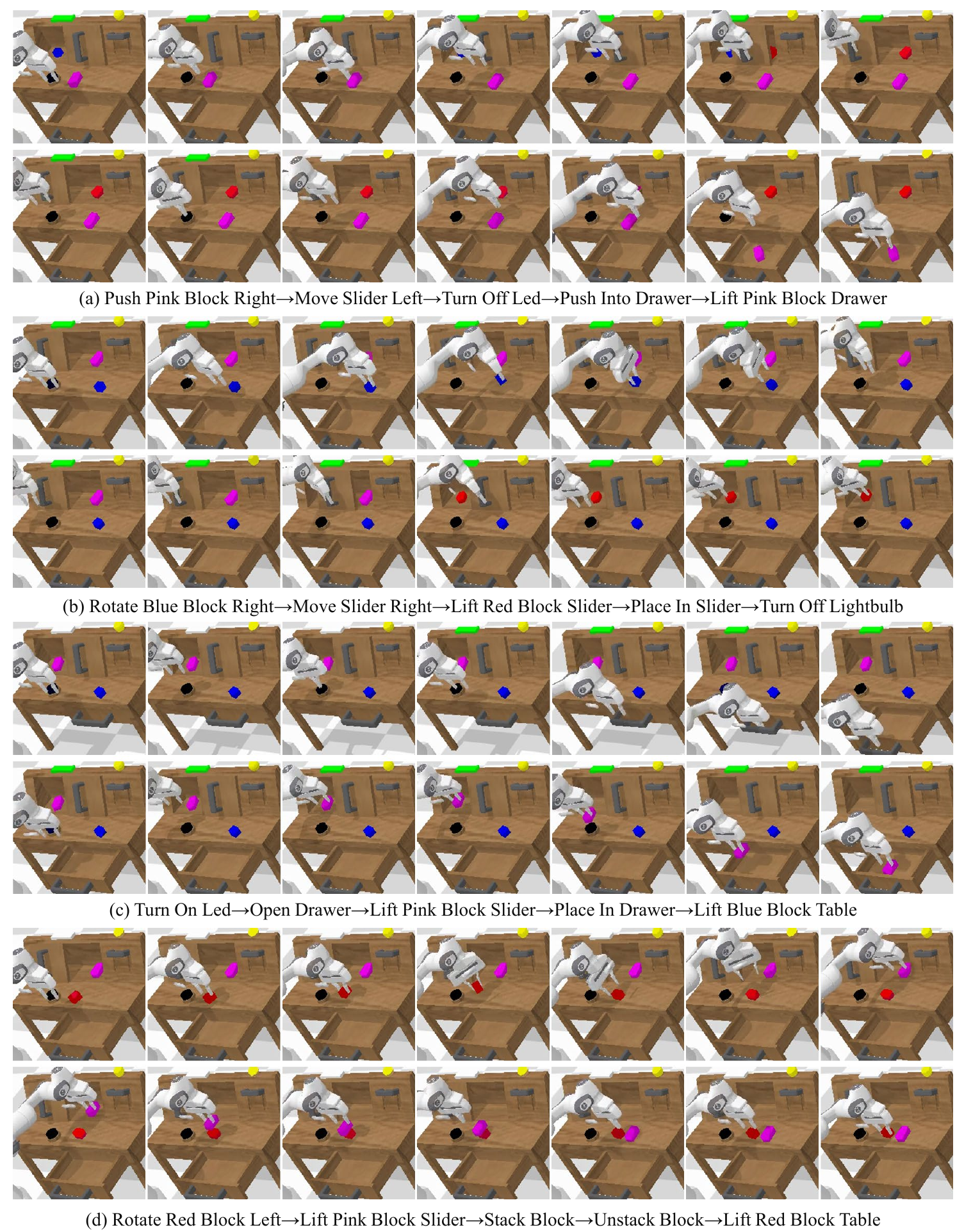}
\caption{Qualitative results on Calvin (1).}
\label{fig:supp_calvin_1}
\end{figure*}

\begin{figure*}[t]
\centering
\includegraphics[width=0.8\linewidth]{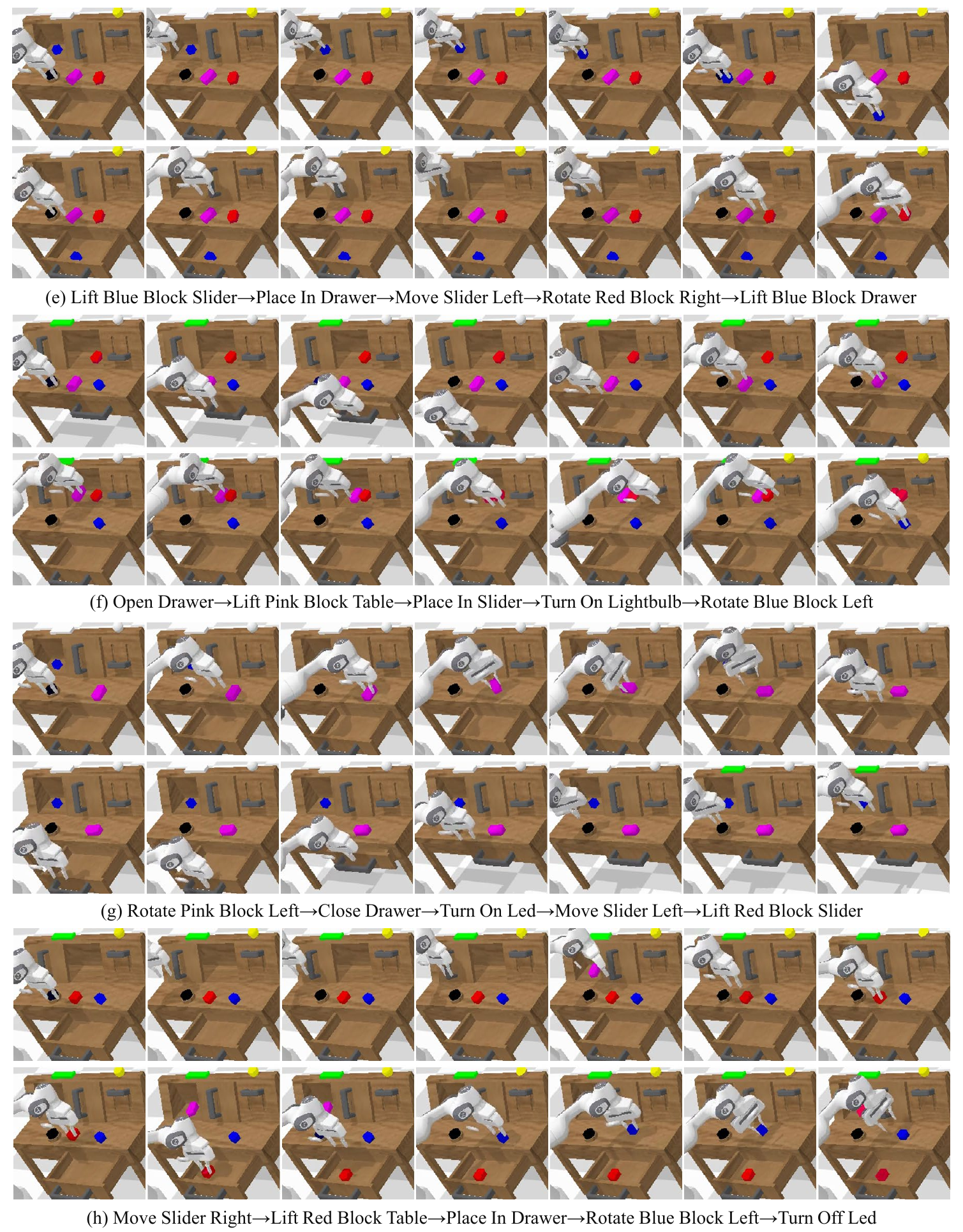}
\caption{Qualitative results on Calvin (2).}
\label{fig:supp_calvin_2}
\end{figure*}

\begin{figure*}[t]
\centering
\includegraphics[width=0.8\linewidth]{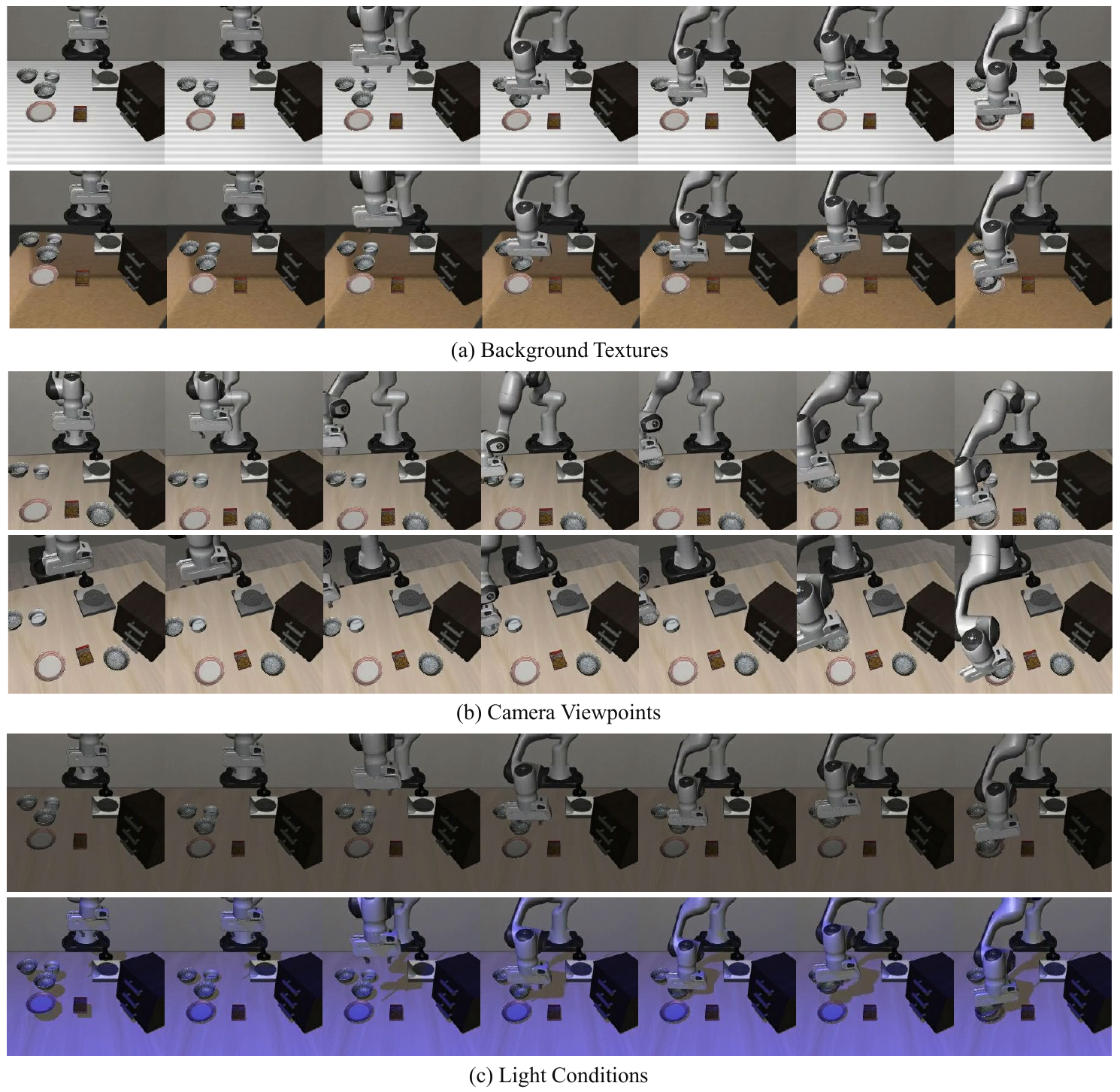}
\caption{Qualitative results on Libero-Plus (1).}
\label{fig:supp_libero_plus_1}
\end{figure*}

\begin{figure*}[t]
\centering
\includegraphics[width=0.8\linewidth]{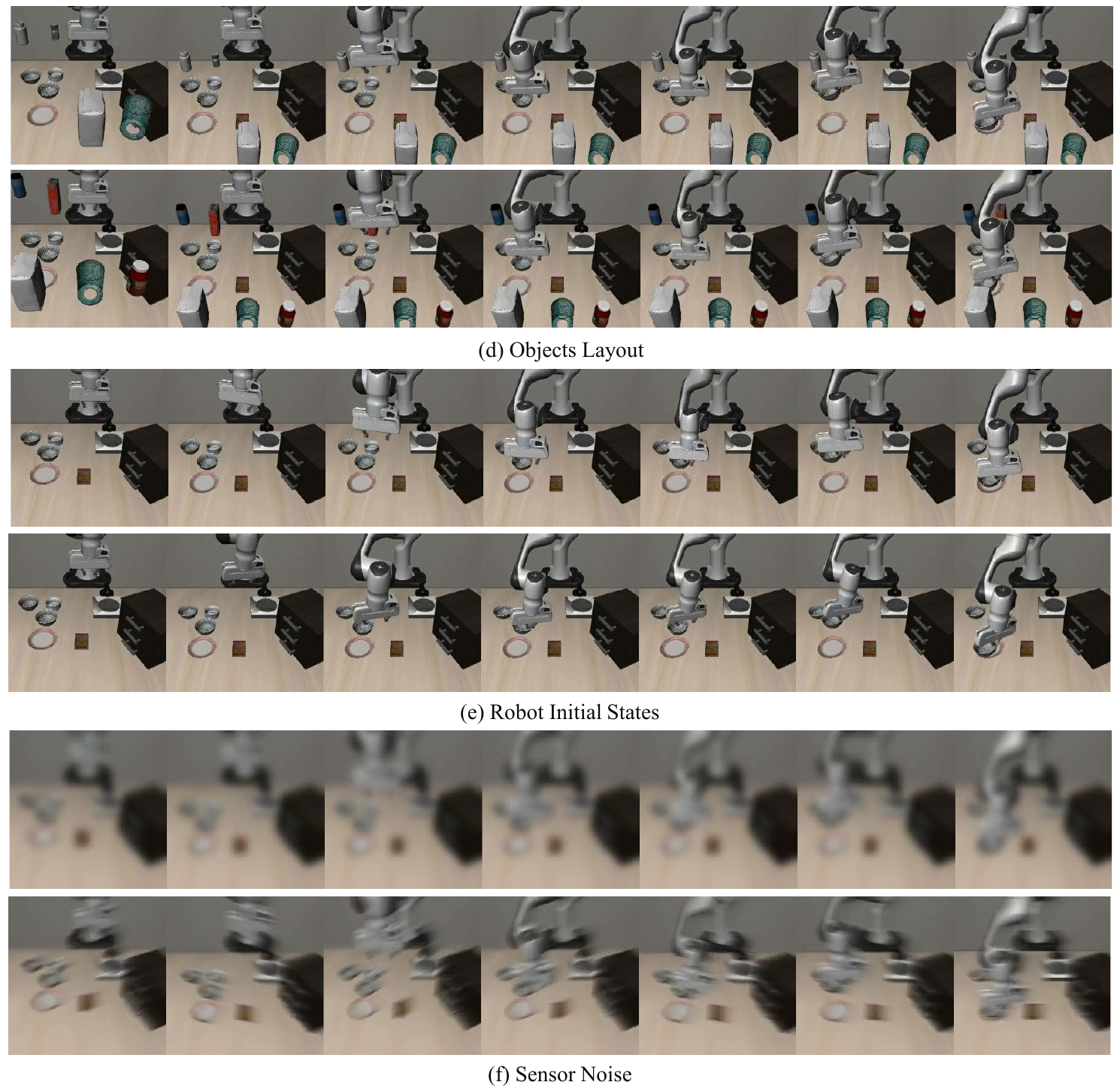}
\caption{Qualitative results on Libero-Plus (2).}
\label{fig:supp_libero_plus_2}
\end{figure*}

\begin{figure*}[t]
\centering
\includegraphics[width=0.8\linewidth]{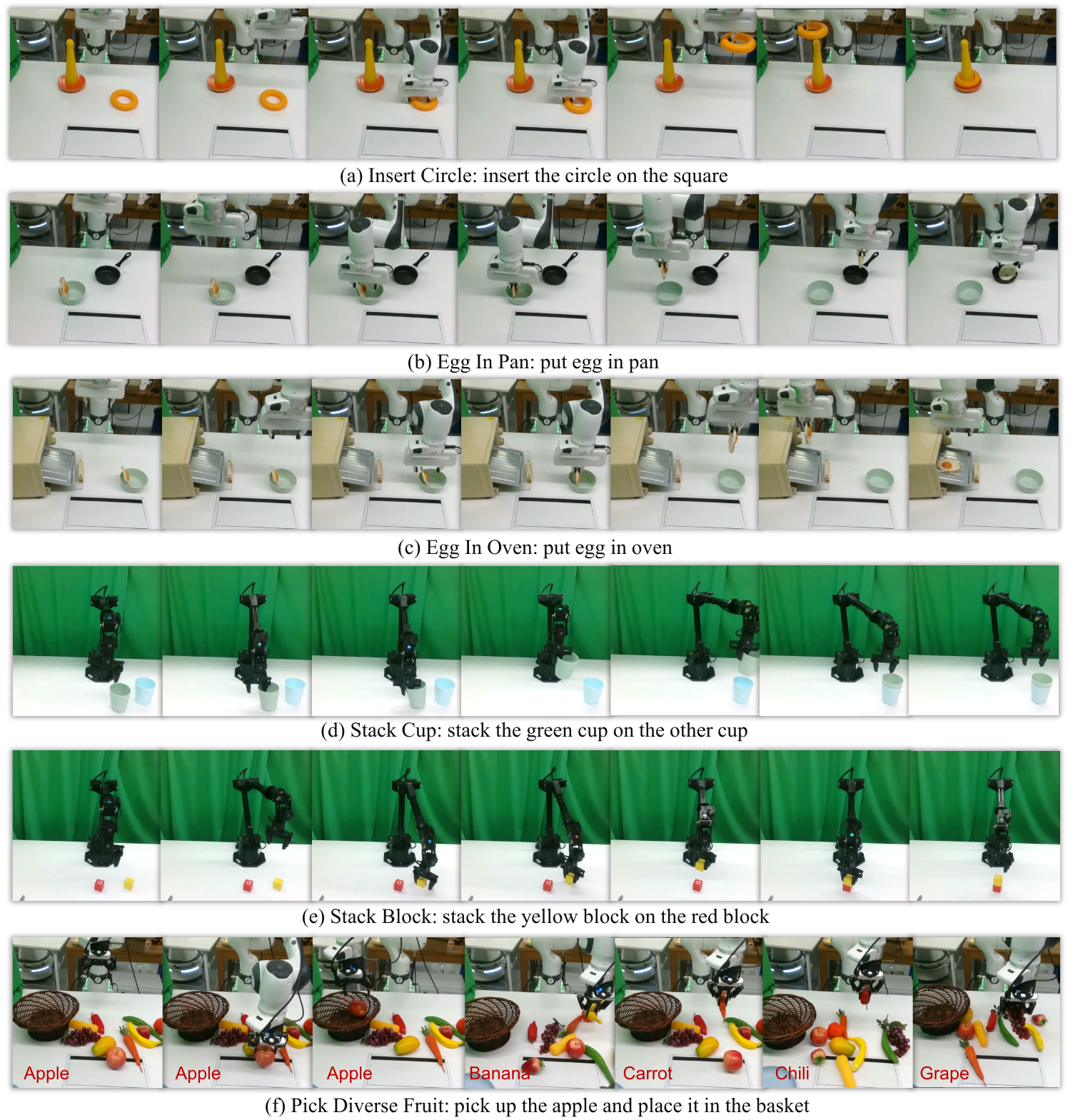}
\caption{Qualitative results on real-world general manipulation tasks.}
\label{fig:supp_real_general}
\end{figure*}

\begin{figure*}[t]
\centering
\includegraphics[width=0.8\linewidth]{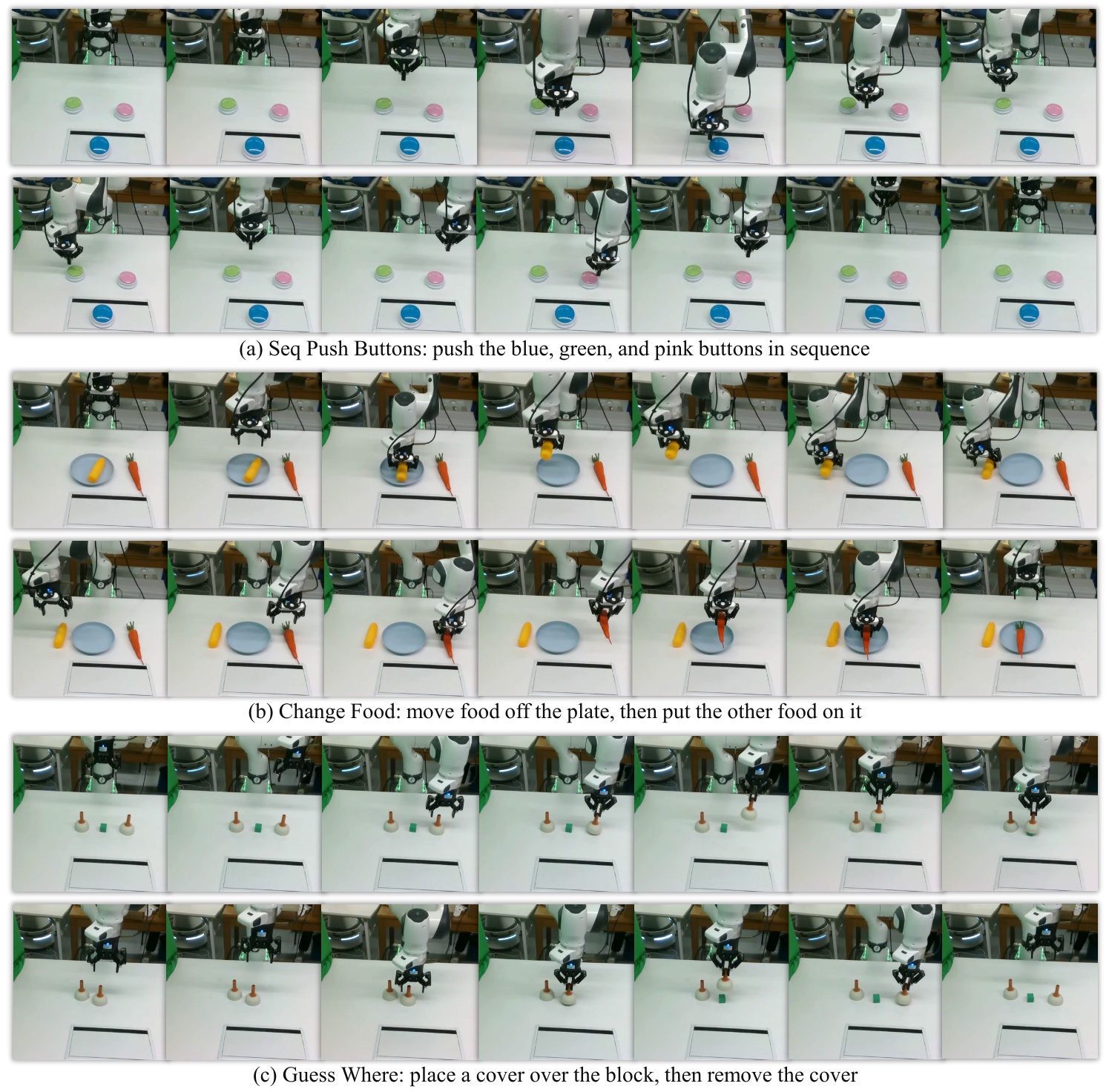}
\caption{Qualitative results on real-world memory-dependent tasks (1).}
\label{fig:supp_real_mem_1}
\end{figure*}

\begin{figure*}[t]
\centering
\includegraphics[width=0.8\linewidth]{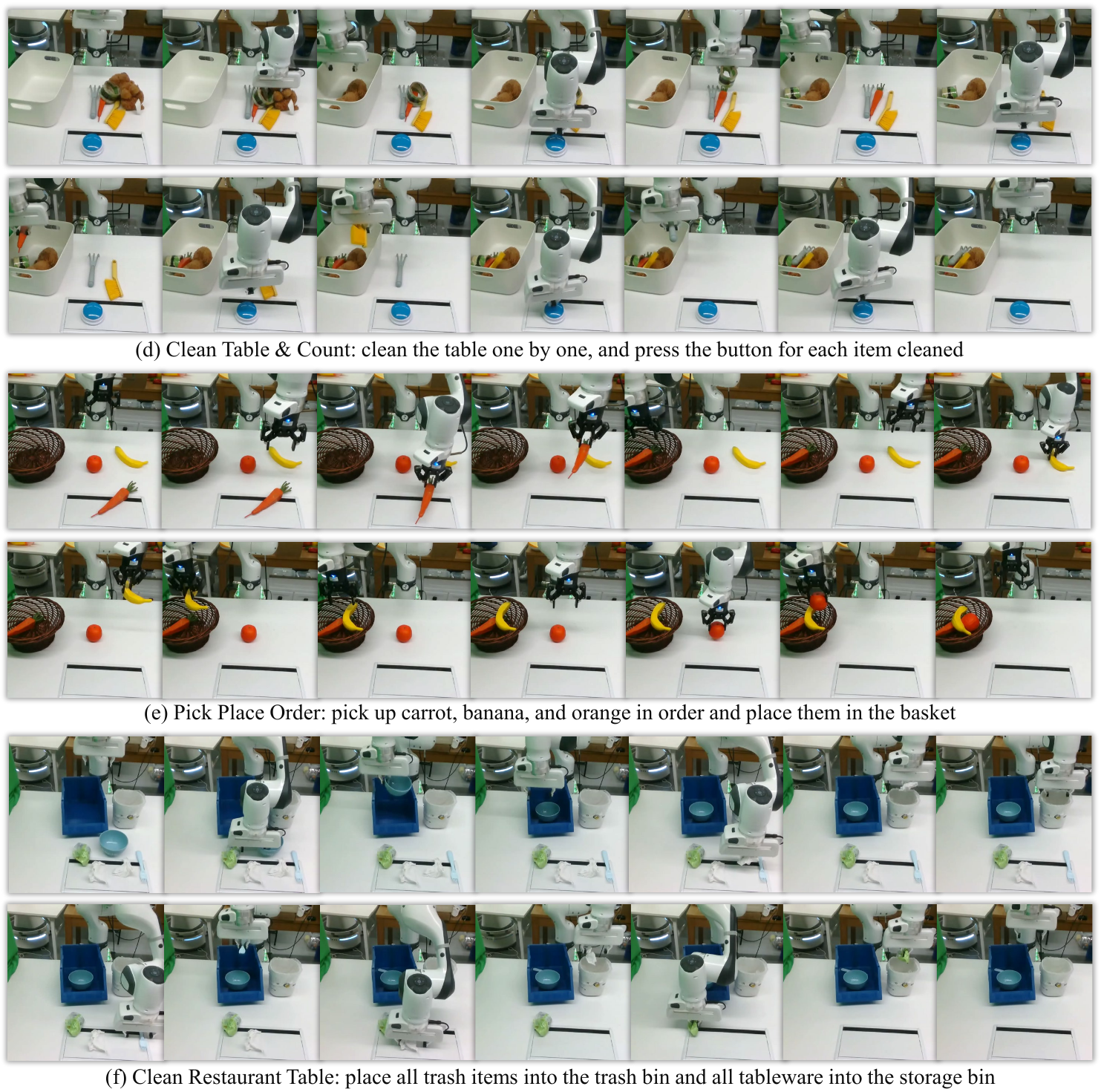}
\caption{Qualitative results on real-world memory-dependent tasks (2).}
\label{fig:supp_real_mem_2}
\end{figure*}

\begin{figure*}[t]
\centering
\includegraphics[width=0.8\linewidth]{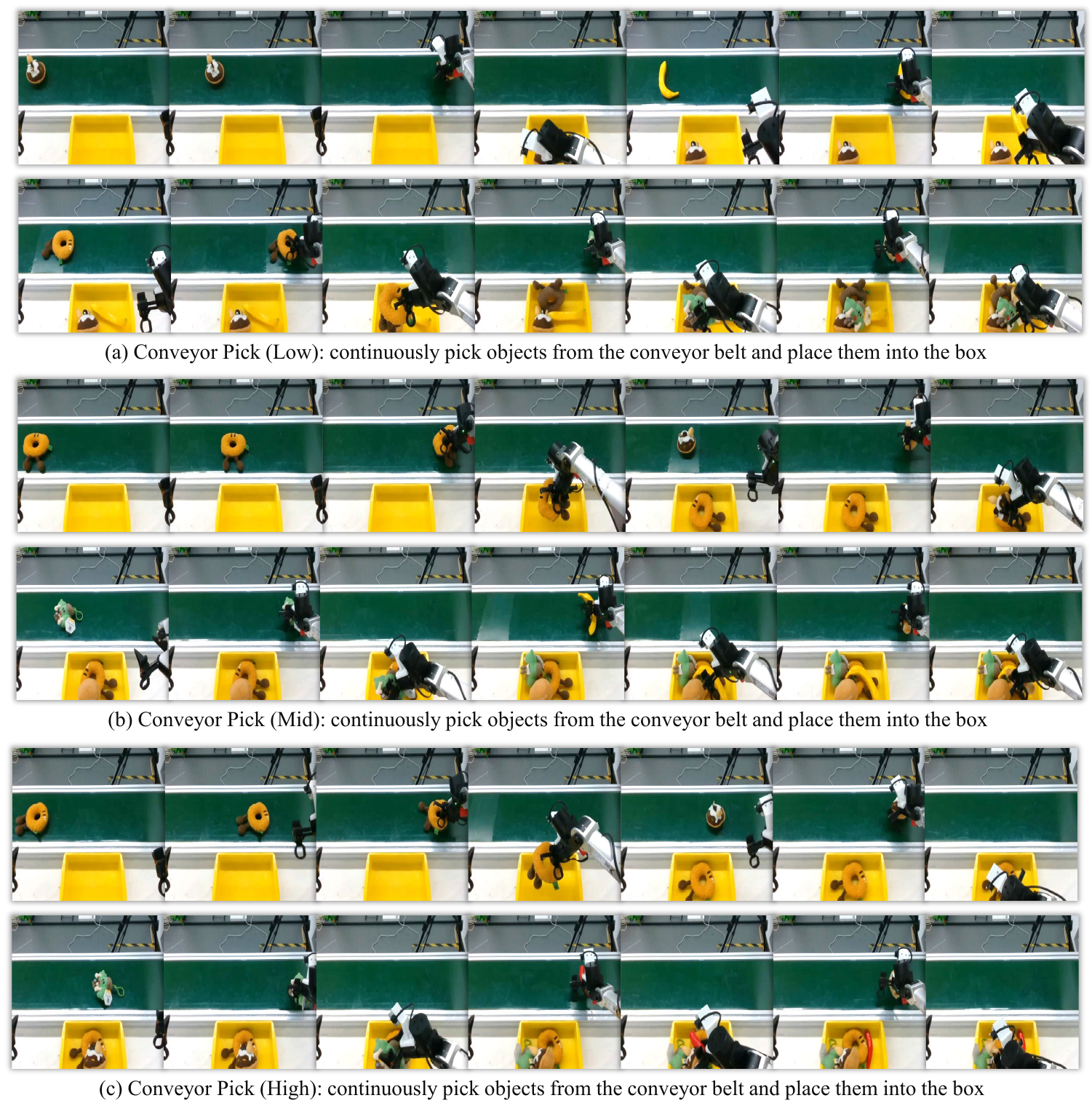}
\caption{Qualitative results on real-world imagination-dependent tasks (1).}
\label{fig:supp_real_imag_1}
\end{figure*}

\begin{figure*}[t]
\centering
\includegraphics[width=0.8\linewidth]{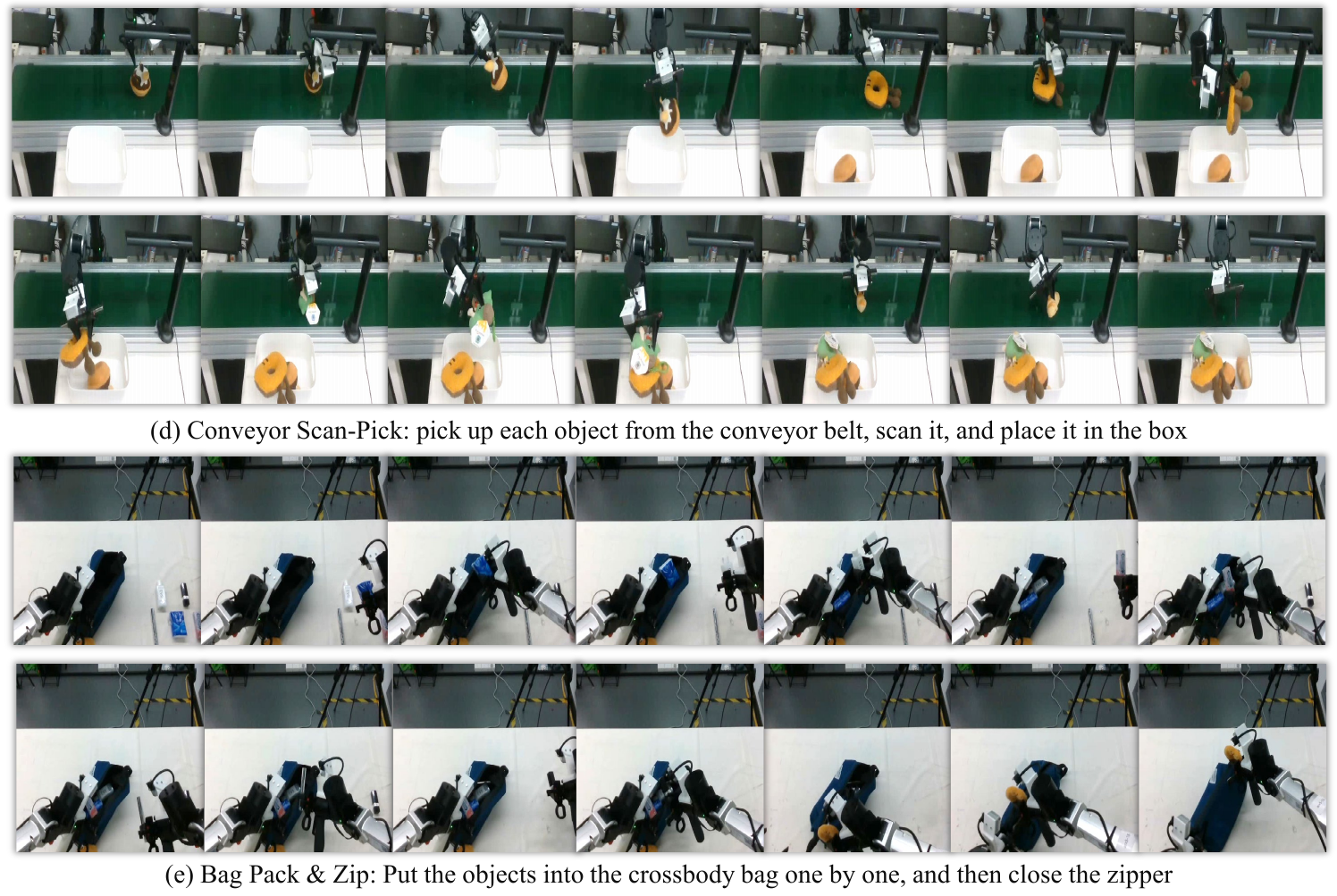}
\caption{Qualitative results on real-world imagination-dependent tasks (2).}
\label{fig:supp_real_imag_2}
\end{figure*}

\begin{figure*}[t]
\centering
\includegraphics[width=0.8\linewidth]{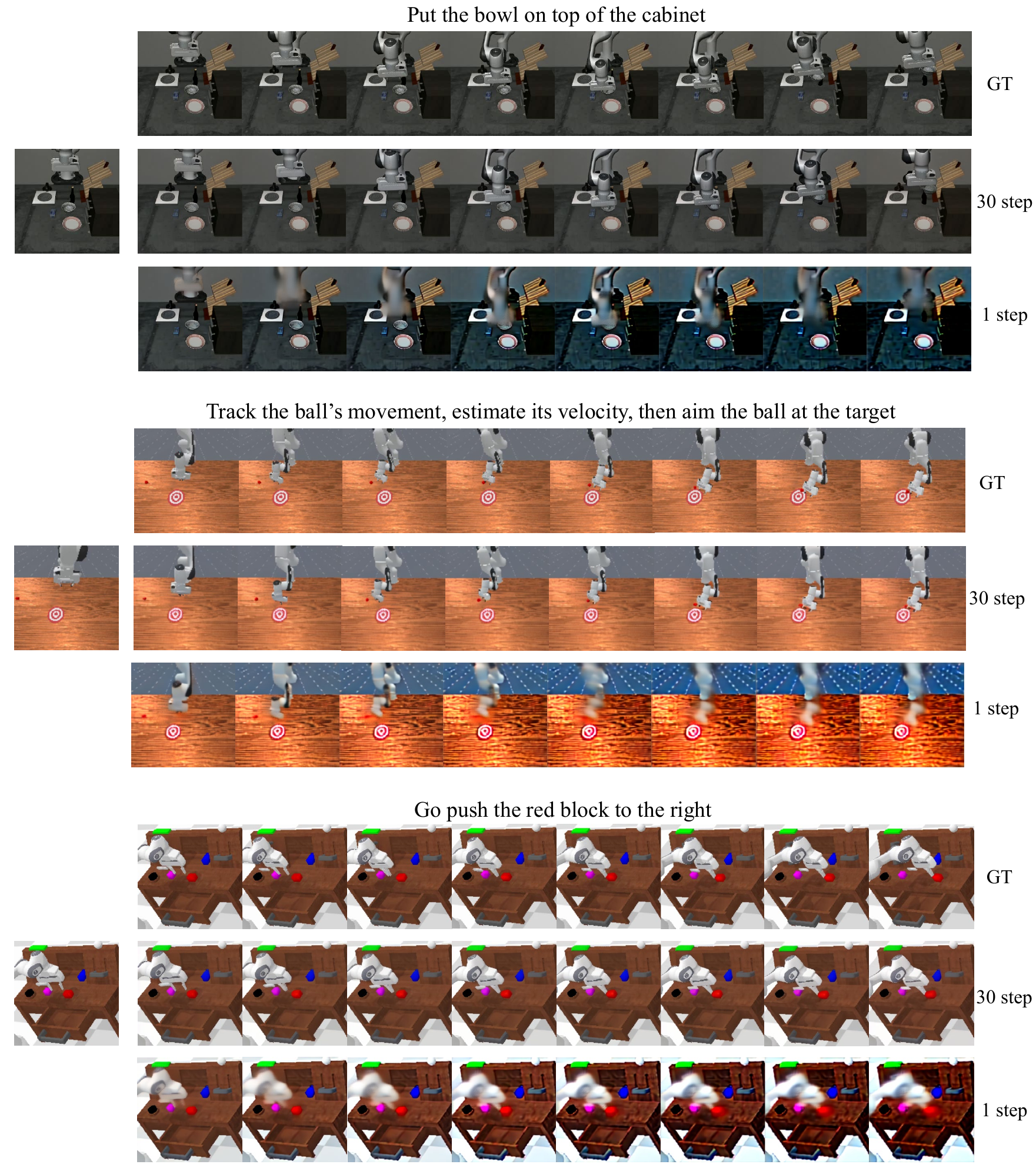}
\caption{Qualitative results of world-model-based future imagination (1).}
\label{fig:supp_world_1}
\end{figure*}

\begin{figure*}[t]
\centering
\includegraphics[width=0.8\linewidth]{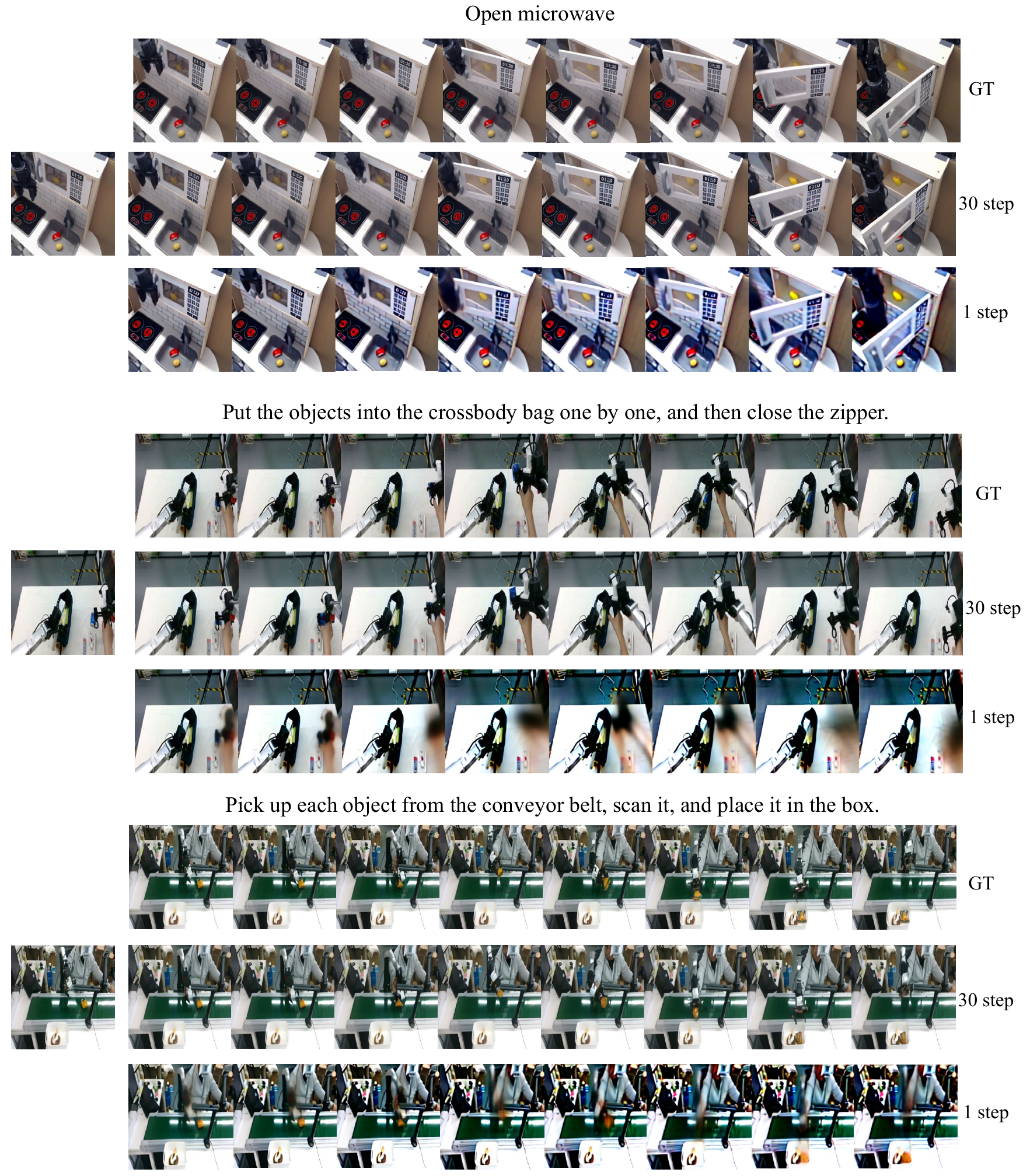}
\caption{Qualitative results of world-model-based future imagination (2).}
\label{fig:supp_world_2}
\end{figure*}

We provide additional qualitative results across simulation and real-world benchmarks. 
Figs.~\ref{fig:supp_libero}, \ref{fig:supp_simpler}, \ref{fig:supp_mikasa}, \ref{fig:supp_calvin_1}, \ref{fig:supp_calvin_2}, \ref{fig:supp_libero_plus_1}, and \ref{fig:supp_libero_plus_2} show results on Libero, SimplerEnv, Mikasa-Robo, Calvin, and Libero-Plus. 
Figs.~\ref{fig:supp_real_general}, \ref{fig:supp_real_mem_1}, \ref{fig:supp_real_mem_2}, \ref{fig:supp_real_imag_1}, and \ref{fig:supp_real_imag_2} show results on real-world general, memory-dependent, and imagination-dependent tasks. 
Figs.~\ref{fig:supp_world_1}, and \ref{fig:supp_world_2} show qualitative results of world-model-based future imagination.

\subsection{Supplementary Videos}
We also provide supplementary videos to better illustrate the behavior of MemoryVLA++ across simulation and real-world tasks. These videos are included in the supplementary materials and are also available on the project website: \url{https://shihao1895.github.io/MemoryVLA-PP-Web}.

\end{document}